\documentclass{article}

\usepackage{PRIMEarxiv}

\usepackage[utf8]{inputenc} 
\usepackage[T1]{fontenc}    
\usepackage{hyperref}       
\usepackage{url}            
\usepackage{booktabs}       
\usepackage{amsfonts}       
\usepackage{nicefrac}       
\usepackage{microtype}      
\usepackage{lipsum}
\usepackage{fancyhdr}       
\usepackage{graphicx}       
\graphicspath{{media/}}     

\usepackage{subfigure}
\usepackage{amsthm,amsmath,amssymb,bm}
\usepackage{mathrsfs}
\usepackage{color}
\usepackage{smileX}
\usepackage{enumitem}
\usepackage{natbib}
\usepackage{algorithm}
\usepackage{algorithmic}

\newtheorem{myDef}{Definition}[section] 
\newtheorem{proposition}{Proposition}[section]
\newtheorem{thm}{Theorem}[section]

\theoremstyle{remark}
\newtheorem{remark}{Remark}[section]

\title{Constrained Policy Optimization with Explicit Behavior Density for Offline Reinforcement Learning}

\author{
  Jing Zhang \\
 HKUST
   \And
  Chi Zhang  \\
  Kuaishou Technology \\
   \And
Wenjia Wang\\
HKUST(GZ) and HKUST
    \And
    Bing-Yi Jing\\
    SUSTech
}

\begin{document}
\maketitle

\begin{abstract}
Due to the inability to interact with the environment, offline reinforcement learning (RL) methods face the challenge of estimating the Out-of-Distribution (OOD) points. Existing methods for addressing this issue either control policy to exclude the OOD action or make the $Q$ function pessimistic. However, these methods can be overly conservative or fail to identify OOD areas accurately. To overcome this problem, we propose a Constrained Policy optimization with Explicit Behavior density (CPED) method that utilizes a flow-GAN model to explicitly estimate the density of behavior policy. By estimating the explicit density, CPED can accurately identify the safe region and enable optimization within the region, resulting in less conservative learning policies.  We further provide theoretical results for both the flow-GAN estimator and performance guarantee for CPED by showing that CPED can find the optimal $Q$-function value. Empirically, CPED outperforms existing alternatives on various standard offline reinforcement learning tasks, yielding higher expected returns.
\end{abstract}

\section{Introduction}

 As a form of active learning, reinforcement learning (RL) has achieved great empirical success in both simulation tasks and industrial applications \cite{Haarnoja2018,Scherrer2015,levine2016end,kalashnikov2018scalable,sutton2018rlintro,li2019rlapplication,Afsar2022}. The great success of RL is largely due to sufficient information exploration and the online training paradigm  that the agent has plenty of interactions with the environment. However, the merits of RL methods are limited when it is difficult or costly to sequentially interact with the environment. In many real-world scenarios such as driverless  or intelligent diagnostics \cite{levine2020offline}, online interaction is unacceptable. Thus, vanilla RL methods may possibly fail under such circumstances.

One recently developed approach to mitigating the constrained online-interaction is offline RL \cite{levine2020offline,Yu2018,Johnson2016}. Motivated by various data-driven techniques, the offline RL leverages prior experiences to train a policy without interacting with the environment. In offline RL, a critical challenge is distribution shift (also called ``extrapolation error`` in literature). Specifically, vanilla RL methods using the Bellman equation and the $Q$-function approximation probably fail to deliver good estimates for Out-of-Distribution (OOD) points. To this end, two types of offline RL solutions have been proposed. One type is the $Q$-function constraint. By adding constraints on the $Q$-function, this type of solution provides a pessimistic estimation of the $Q$-function and prevents the $Q$-value from being too large. 
The constraints include ensembling multiple $Q$-functions \cite{Agarwal2020,an2021uncertainty}, penalizing $Q$-function with high uncertainty \cite{Kumar2020,wu2021uncertainty}, among others. Recently, implicit Q-learning \cite{kostrikov2021offline} and mildly conservative Q-learning \cite{lyu2022mildly} are suggested. However, the uncertainty of the $Q$-function or a tight lower bound of $Q$-value is difficult to estimate. Thus the performance is not guaranteed neither theoretically nor practically. 

Another type of solution is policy control, where the learning policy is controlled over the support of the behavior policy. Thus the OOD points will not be visited and the distribution shift problem is alleviated \cite{Kumar2019,wu2019behavior,Fujimoto2019}.  
It is noted that previous policy control methods limit the safe areas within the close neighborhoods of the training dataset by distribution distance metrics such as KL divergence, maximum mean discrepancy (MMD), and Wasserstein Distance \cite{jaques2019way,Kumar2019,wu2019behavior}. Such policy control approaches often lead to an over-conservative and suboptimal learning policy, and the optimal policy value may not be reached. 
For example, if some state-action points are within the training distribution but unobserved in the training dataset, they may be considered OOD points in previous policy control approaches. Nevertheless,these points are safe when the estimated $Q$-function can well generalize on this area \cite{an2021uncertainty,Degrave2022}, and the optimal policy likely lies in the unseen area of the data space. In this sense, a global estimator of the conditional action data space is required to define the safe areas and examine whether an unseen point is safe. To this end, \citet{wu2022supported} imposes the density-based constraints and proposes the SPOT method. However, the density estimator in SPOT approximates the lower bound of the behavior policy density, and policy control methods with more accurate density estimators are worth further discussing. As in the field of policy control methods for offline RL problem, the idea of ensuring the consistency of the support of learned policy and that of the behavior policy is considered the most desirable approach to tackle distribution shift. Therefore, a direct estimation density of the behavior policy is desired for guarantee support consistency.

Recently, generative models have shown strong capabilities in diverse fields, in which variational autoencoder (VAE) and generative adversarial network (GAN) \cite{Goodfellow14,Arjovsky17,li2017mmd,ho2020denoising} are popular tools utilized for sampling and other related applications. In RL areas, both VAE and GAN have been introduced in RL methods \cite{Fujimoto2019,lyu2022mildly,Yarats2021vae,Doan2018ganqlearning}. However, as random sample generators, VAE and GAN cannot be directly applied to estimate behavior densities in offline RL, nor can they provide accurate density estimates. 
\citet{wang2022diffusion,yang2022behavior,chen2022offline,ghasemipour2021emaq} have shown that VAE has shortcomings in estimating complex distributions, especially multimodal distributions or complex behavior policy distributions, and discuss that typically-used VAEs may not align well with the behavior dataset. As VAE is prone to cover mode, this phenomenon diminishes the VAE's capacity to accurately estimate data distributions, particularly in scenarios involving multimodal data distributions, such as behavior policy distributions. Compared to VAE, GAN has more advantages in learning complex data distributions. Recently, various GAN variants have been suggested. Among them, the Flow-GAN \cite{Grover2018} model combines the normalizing flow model \cite{Dinh2015,Dinh2017} and GAN, and provides an explicit density estimator. The Flow-GAN model is further renowned for its robustness and tuning flexibility and has garnered increasing attention.  

\begin{figure}[h!]
\vskip 0.2in
\begin{center}
\centerline{\includegraphics[width=\columnwidth]{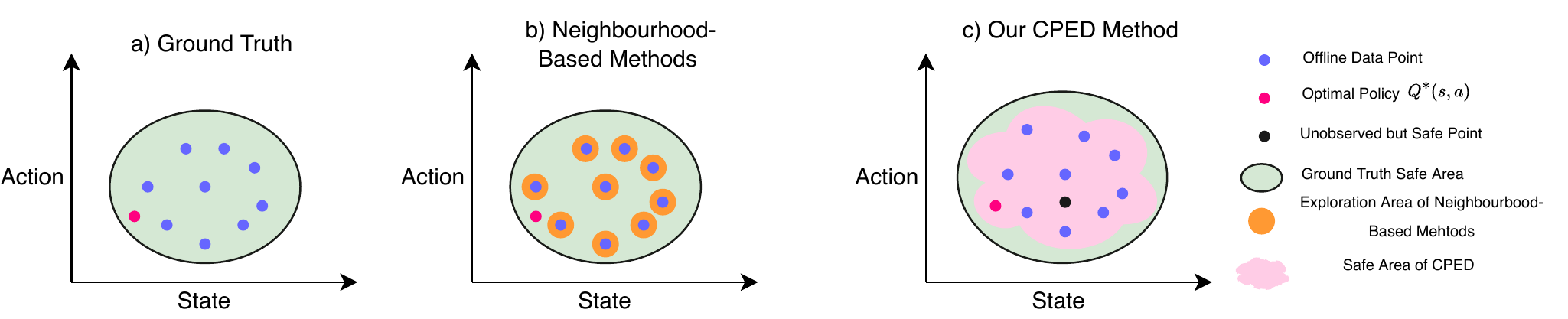}}
\caption{ (a): The ground truth safe area in offline RL optimization, and the updates of policies and $Q$-functions are done within the green area. The blue points are collected behavior data $\cD$, and the red point denotes the optimal policy given the states. (b): In previous approaches, the exploration of the policy takes place in a small neighborhood of points in $\cD$ (the orange circles). (c): The CPED relaxes the exploration area and constructs the feasible region (pink areas), which includes the unobserved but safe points (black point).}
\label{fig1}
\end{center}
\vskip -0.3in
\end{figure}

In this work, we propose a novel approach called Constrained Policy optimization with Explicit behavior Density (CPED\footnote{Code available at https://github.com/evalarzj/cped}) for offline RL. The CPED suggests a strong density-based constraint and employs an explicit and accurate density estimator to identify the safe areas. 
Specifically, CPED utilizes the Flow-GAN model as the density estimator, whose accuracy is theoretically guaranteed. By leveraging this capability, CPED can accurately identify the feasible region, which includes both observed and unobserved but safe points (illustrated in Figure \ref{fig1} c). By expanding the feasible region and enabling reasonable exploration within it, CPED exhibits reduced risk-aversion and demonstrates an enhanced ability to generate superior policies.

To summarize, the merits of the proposed CPED include the following. 
\begin{itemize}
\item  By introducing the Flow-GAN model, the CPED directly tackles the density estimation problem and proposes a novel policy control scheme for solving offline RL tasks. 
\item  The CPED does not behave over-conservatively and allows exploration in the safe region, especially high-density areas, to find the optimal policy. 
\item  Practically, the effectiveness of CPED is fully demonstrated empirically. Extensive experiments on competitive tasks show that CPED substantially outperforms state-of-the-art methods. 
\item  From the theoretical perspective, we show that GAN with a hybrid loss can accurately estimate density, which itself contributes to the GAN community. Furthermore, we prove the policy learned from CPED can access the optimal value function, theoretically ensuring the efficiency of CPED. 
\end{itemize}

\section{Related Work}

\textbf{Value Function Constraint}. Taking OOD actions can lead to overestimated value function and introduce bootstrap error. Value function constraint methods aim to construct pessimistic value functions or reduce the uncertainty of the value function to avoid OOD actions. Following \cite{Agarwal2020,an2021uncertainty,Fujimoto2018}, a pessimistic value function can be obtained by ensembling multiple value functions. CQL \cite{Kumar2020} proposes a regularization term so that the estimated $Q$-function low-bounds its true value. TQC \cite{kuznetsov2020controlling} follows the distributional RL method and learns the percentile of value function distribution. IQL \cite{kostrikov2021offline} combines expectile regression with exponential weighted advantage function to avoid extreme value function. MCQ \cite{lyu2022mildly} tries to use a special bellman operator to separate the estimation of the target value function for the in-distribution point and OOD point. The value function shows great uncertainty when taking OOD points. UWAC \cite{wu2021uncertainty} estimates the uncertainty of the value function using the dropout technique. \citet{urpi2021risk} applies the tail risk measurement for the $Q$-function, in which the value function with high uncertainty is penalized, and the learned value function is forced to be pessimistic. However, finding the lower bound of the value function or estimating the uncertainty is not easy. The error from estimating the value function and finding the lower bound hinders the performance of value function constraint methods. Recently, a series of in-sample learning methods \cite{garg2023extreme,xiao2023sample,zhang2022sample,xu2023offline,hansen2023idql} have been proposed to optimize the $Q$-value function or the policy value function $V$ solely utilizing in-sample data. As a result, the extrapolation error, induced by out-of-distribution (OOD) data points in the $Q$-value function, can be mitigated.

\textbf{Policy Control Method}. To prevent the learning of policies that take out-of-distribution (OOD) actions, the policy control method incorporates a penalty term during policy learning and restricts the distribution distance between the learned policy and the behavior policy. The BEAR method \cite{Kumar2019} uses the MMD distance to constrain the difference between the behavior policy and the learned policy. \cite{wu2019behavior, nair2020awac,
jaques2019way,fujimoto2021minimalist,li2022doge} measures the distribution shift between learning policy and behavior policy with KL divergence and Wasserstein distance. DOGE \cite{li2022doge} learns a distance function to control the learning policy staying close to the behavior policy. Besides the explicit distance-based policy control methods, some implicit control methods \cite{Fujimoto2019,zhou2021plas,wu2022supported}, which generate the learning policy action based on the estimated conditional probability distributions, are suggested. However, explicit distance control methods suffer from the generalization problem since the distance is estimated from the limited training samples. On the other hand, implicit control methods tend to generate over-conservative policies, where the generated actions are too close to the behavior policy. Consequently, the learned policy by these methods discourages exploration and remains conservative.

\textbf{Behavior Policy Density Estimation}
In offline RL setting, determining the action space $\mathscr{A}$ under given states is crucial. This is equivalent to learning the distribution of the behavior policy. However, traditional density estimation methods \cite{hathaway1985,wang2015,friedman2001elements,Hall1981,wahba1981,Donoho96} 
 often fail in this scenario of density estimation due to the curse of high dimensionality. Recently, deep neural network-based density estimation methods have proven successful in learning the distribution of the behavior policy. Conditional variational autoencoders (CVAE) have been utilized as both behavior policy samplers \cite{Fujimoto2019,Kumar2019,zhou2021plas,lyu2022mildly,chen2022latent} and density estimators \cite{wu2022supported}. Generative adversarial networks (GAN) are also introduced in offline RL scenario \cite{yang2022regularized,yang2022regularizing}. \citet{singh2020parrot} applies the generative normalized flow model to learn an invertible map from the noisy prior to the behavior policy action space and do behavior policy sampling by the trained model. EMaQ \cite{ghasemipour2021emaq} uses the Autoregressive Generative Model instead of CVAE as a sampler. Additionally, the diffusion model has been applied to estimate the density of the behavior policy in both model-free RL \cite{chen2022offline} and model-based RL \cite{janner2022planning}. Specifically, SPOT \cite{wu2022supported} approaches the lower bound of the behavior density with ELBO derived from CVAE. However, to the best of our knowledge, most of the aforementioned methods either estimate the behavior density implicitly or lack the ability to provide accurate estimations. In our proposed method, the CPED finds an exact density function using the Flow model \cite{Dinh2015,Dinh2017} and GAN \cite{Goodfellow14,Arjovsky17,li2017mmd,gulrajani2017improved,radford2015unsupervised}.

\section{Probability Controlled Offline RL Framework}

\subsection{Offline RL Problem Settings}
In RL setting, the dynamic system is described in terms of a Markov decision process (MDP), which can be defined by a set of tuples $\{\mathcal{S},\mathcal{A},T,p_0,r,\gamma\}$. Here, $\mathcal{S}$ and $\mathcal{A}$ denote the state space and  action space, respectively; $T$ is a conditional transition kernel, and $T(s'|s,a)$ is the conditional transition probability between states, describing the dynamics of the entire system; $p_0$ is the distribution of the initial states; $r(s,a)$ is a deterministic reward function; and $\gamma\in (0,1)$ is a scalar discount factor.

The goal of RL is to find a distribution of actions $\pi(a|s)$ (named as policy) that yields the highest returns within a trajectory, i.e., maximizing the expected cumulative discounted returns. We denote this expected discounted return starting from $(s, a)$ under policy $\pi$ as $Q^{\pi}(s,a)$. Under the Q-learning framework, the optimal expected discounted return $Q^*(s,a)$ can be obtained by minimizing the $L_2$ norm of Bellman residuals:
\begin{align}\label{eq_bellmanR}
    \mathbb{E}_{s'\sim T(s'|s,a),a'\sim \pi(a|s)}[Q(s,a)-(\cB Q)(s,a)]^2,
\end{align}
where $\cB$ is the Bellman operator defined as
\begin{align*}
    (\cB Q)(s,a):=r(s,a)+\gamma \mathbb{E}_{s''\sim T(s'|s,a)}[\max_{a'}Q(s',a')].
\end{align*}

In the context of deep RL, various solutions are provided to estimate the $Q$-function \cite{Mnih2013,Hasselt2016,wang2016}. When the action space is continuous, the learning process is divided into actor ($Q$ value function) training and critic (learning policy) training \cite{Sutton1998}. The critic network is learned from Eq.\ref{eq_bellmanR} without ``maximum'' in the Bellman operator. The actor is formulated with various forms \cite{Silver2014,Lillicrap2015,Schulman2017,Haarnoja2018} such as maximizing the $Q$-value (DDPG), the weighted log of action density (policy gradient), and clipped weighted action density ratio (PPO).

In the offline RL, the learning policy cannot interact with the dynamic system yet only learns from a training dataset $\cD(s,a,s',r)$. Dataset $\cD$ contains many triples $(s,a,s',r)$ that can be viewed as independent. 
Assume that the dataset $\cD$ is generated by a behavior policy $\pi_{\beta}$. We need to redefine the MDP corresponding to $\cD$ under the offline RL settings.

\begin{myDef}[Offline MDP]\label{def1}
Given a dataset $\cD$ with continuous state and action space. Let $\pi_{\beta}$ denote the behavior policy that generated $\cD$. The MDP of dataset $\cD$ is defined by $\mathcal{M}_{\cD}:=\{\mathscr{S},\mathscr{A},T_{\cD},p_{0_{\cD}},r,\gamma\}$, where $\mathscr{S}=\{\mathcal{S},\mathbb{P}_{\cD}\}$ is the measured state space related to $\cD$ with probability measure $\mathbb{P}_{\cD}$, $\mathbb{P}_{\cD}$ is the marginal probability of each state in state space of $\cD$, $\mathscr{A}=\{\mathcal{A},\pi_{\beta}\}$ is the measured action space related to $\cD$ with conditional probability measure $\pi_{\beta}(a|s)$, $T_{\cD}(s'|s,a)$ is the transition probability between states, describing the dynamics of $\cD$, and $p_{0_{\cD}}$ is the distribution of the initial states of $\cD$.
\end{myDef}

In offline RL, we can obtain the optimal $Q$-function by minimizing Eq.\ref{eq_bellmanR}. However, due to the distribution shift problem \cite{Kumar2019,levine2020offline}, the Bellman residuals are biased during the training phase. Specifically, at $k$-th iteration, since $\pi_k$ is trained to maximize the $Q$-function, it is possible that $\pi_k$ visits some OOD points with high estimated $Q$-function values. Nevertheless, we cannot evaluate the rewards of the OOD points under offline scenarios. Consequently the Bellman residuals at the OOD points may dominate the loss, and the learned $Q$-function value is mislead. 
Therefore, the optimization of offline RL must occur in the bounded safe state-action space ${\mathscr{S} \times \mathscr{A}}$, such that the estimated $Q$-function is trustworthy.

\subsection{Behavior Policy Estimation by GAN with Specific Density}

In offline RL, the dataset $\cD$ is generated by the behavior policy, whose density can be estimated via inverse RL \cite{ng2000algorithms}. One widely used approach in inverse RL, maximum entropy inverse RL (MaxEnt IRL, \cite{ziebart2008maximum}), models the density via 
a Boltzmann distribution 
\begin{equation}\label{eq1}
p_{\theta}(\tau)=Z^{-1}\exp(-c_{\theta}(\tau)),
\end{equation}
where the energy is given by the cost function $c_{\theta}(\tau)=\sum_{t=0}^H c_{\theta}(s_t,a_t)$, $\tau=(s_0,a_0,s_1,a_1,...,s_H,a_H)$ is a trajectory, and $Z$ is a scale factor. The cost function
$c_{\theta}(\tau)=\sum_{t=0}^H c_{\theta}(s_t,a_t)$ is a parameterized cost function that can be learned by maximizing the likelihood of the trajectories in $\cD$. And the scale factor $Z$ is the sum of exponential cost $\exp(-c_{\theta}(\tau))$ of all the trajectories that belong to the product space of the state and action of $\cD$. However, estimating $Z$ is difficult because it is scarcely possible to traverse the entire action space $\mathscr{A}$. An alternative way to estimate $Z$ is via guided cost learning \cite{Finn2016}, where it has been shown that learning $p_{\theta}(\tau)$ is equivalent to training the dataset $\cD$ with a GAN \cite{Finn2016,finn2016connection,Ho2016,ho2016generative}. 

Originally, the GAN is a random sample generator for high-dimensional data, and it consists of two models: a generator $G$ and a discriminator $D$. The generator is trained to generate random samples that are close to those drawn from the underlying true distribution, and the discriminator tries to classify whether the input is generated sample or an actual one from the training dataset. The GAN cannot provide an explicit density estimation, which is an indispensable ingredient when controlling the probability of actions and specifying an safe area. Thus, we cannot directly apply the MaxEnt IRL to our CPED. To overcome the difficulty, we follow the approach in \cite{Grover2018} and consider GAN with flow model as a generator (we call if Flow-GAN hereafter as in \cite{Grover2018}).

Flow-based generative models have shown their power in the fields of image processing and natural language processing \cite{Dinh2015,Dinh2017}. The normalizing flow model allows us to start from a simple prior noise distribution and obtain an \textit{explicit} estimate of the density function after a series of non-linear invertible transformations.

Specifically, let $f_\theta: \mathbb{R}^A \mapsto \mathbb{R}^A$ be a nonlinear invertible transformation, where $A$ is the dimension of action space $\mathscr{A}$, and $\theta$ is parameter. By change of variables, the probability density on $\cD$ can be obtained by 
\begin{equation}\label{eq3}
p_\theta(\tau)=p_H(f(\tau))\left|{\rm det}\frac{\partial f_\theta(\tau)}{\partial \tau}\right|,
\end{equation}
where $\tau \in \cD$ is a trajectory. Since $f_\theta$ is invertible, a generator can be constructed by $G_\theta=f_\theta^{-1}$. In practice, both generator and discriminator in GAN are modeled as deep neural networks. Two specific neural network structures, NICE \cite{Dinh2015} and Real\_NVP \cite{Dinh2017}, are proposed to ensure that $G_\theta$ (or $f_\theta$) is invertible. The Jacobian matrices of the mapping functions $f$ constructed by these two methods are both triangular matrices, guaranteeing efficient computation of $|{\rm det}\frac{\partial f(\tau)}{\partial \tau}|$.

Directly training a Flow-GAN by original loss function in GAN may lead to poor log-likelihoods, and GAN is prone to mode collapse and producing less diverse distributions \cite{Grover2018}. Thus, following the approach in \cite{Grover2018}, we adopt a hybrid loss function containing a maximum likelihood loss and a GAN structure loss as the optimization objective for estimating the probability measure of behavior policy:
\begin{equation}\label{eq4}
\min_{\theta}\max_{\phi}\mathcal{L_{G}}(G_{\theta},D_{\phi})-\lambda\mathbb{E}_{\tau \sim P_{data}}[\log(p_{\theta}(\tau))]
\end{equation}
where $\mathcal{L_G}(G_{\theta}, D_{\phi})$ can be any min-max loss function of GAN, $D_\phi$ is a discriminator indexed by parameter $\phi$, and $\lambda>0$ is a hyperparameter balancing the adversary learning process and the maximum likelihood process. 

\begin{remark}
The reason for not simply using MLE is that MLE tends to cover all the modes of distribution (even though some modes have small probabilities), and MLE is not robust against model misspecification \cite{white1982maximum}, which often occurs in high dimensional scenarios.
\end{remark}
In the following proposition, we show that  the estimated density of the behavior policy using the hybrid loss Eq.\ref{eq4} is  equivalent to that using MaxEnt IRL. The proof of Proposition \ref{prop1} is provided in Appendix \ref{appb}.

\begin{proposition}\label{prop1}
For offline dataset $\cD$ generated by behavior policy $\pi_{\beta}$, the learned likelihood function $L^{\pi_{\beta}}$, using GAN with hybrid loss in Eq.\ref{eq4} is equivalent to that trained by MaxEnt IRL. If the generator of GAN can give a specific likelihood function $p_{\theta}^G(\tau)$, then 
\begin{equation}\label{eq2}
L_{\theta}^{\pi_{\beta}}(\tau) = CZ^{-1}\exp(-c_{\theta}(\tau)) \propto p_{\theta}^G(\tau)
\end{equation}
where $C$ is a constant related to $\cD$.
\end{proposition}

\subsection{Constrained Policy Optimization with Explicit Behavior Density Algorithm (CPED)}
\label{CPED}

In this work, we propose an algorithm for probabilistically controlling the learned policy, CPED. In CPED, we first utilize the Flow-GAN and obtain a probability estimate of the trajectory $p_{\theta}(\tau)$. The optimize objective follows the hybrid loss in Eq.\ref{eq4}:
\begin{equation}\label{eq5}
    \min_{\theta}\max_{\phi}\mathcal{L_G}(G_{\theta},D_{\phi})-\lambda\mathbb{E}_{(s,a) \sim D}[\log(L_{\theta}^{\pi_{\beta}}(s, a)))].
\end{equation}
Since the episode $\tau$ is a sequential collection of states $s$ and actions $a$, we rewrite the input of $L_{\theta}^{\pi_{\beta}}(\cdot)$ with $(s, a)$ instead of $\tau$ in Eq.\ref{eq5} and afterwards. From Proposition \ref{prop1}, the solution to 
Eq.\ref{eq5} is 
equivalent to the behavior estimator obtained from MaxEnt IRL, thus providing an effective estimation of the behavior policy.

Following the learning paradigm in classic RL methods, the policy learning process is divided into two parts: the actor training and critic training. With the estimated function $L_{\theta}^{\pi_{\beta}}(\cdot)$ approximating the density of behavior policy, we can naturally propose a density-based constraint in policy learning, and obtain the safe area according to the estimated density. The points with low density values are excluded and the updated policy cannot are exceed the safe region. The actor learning is formulated as: 
\begin{align}\label{eq7}
    & \max_{\psi}~\mathbb{E}_{s \sim \mathbb{P}_{\cD}, a \sim \pi_{\psi}(\cdot|s), (s,a) \in \tilde{\cS}\times \tilde{\cA}}[Q_{\eta}(s,a)],
\end{align}
where $\pi_{\psi}(\cdot|s)$ is the learning policy, and $\tilde{\cS}\times \tilde{\cA}$ is an estimate of the bounded safe area ${\mathscr{S} \times \mathscr{A}}$ in Definition \ref{def1}. 

There are many approaches for specifying the safe area $\tilde{\cS}\times \tilde{\cA}$. 
\citet{Kumar2019} bounds the expected MMD distance between the learning policy and the behavior policy within a predetermined threshold, while \citet{wu2022supported} constrains the lower bound of the estimated behavior density  to identify the safe region.
  In this work, since we can have a direct density estimation of $L_{\theta}^{\pi_{\beta}}(s,a)$, we can utilize it and specify the safe region as the region with not too small density. Specifically, we set 
$\tilde{\cS}\times \tilde{\cA} = \{(s,a)\in \cS\times \cA : -\log L_{\theta}^{\pi_{\beta}}(s,a)<\epsilon\}$, where $\epsilon$ is the threshold\footnote{Eq.\ref{eq7} is a constrained optimization problem, and we are indeed using Lagrangian techniques to solve the constraint problem. During the practical optimization, the constraint on $\tilde{\cS}\times \tilde{\cA} = \{(s,a)\in \cS\times \cA : -\log L_{\theta}^{\pi_{\beta}}(s,a)<\epsilon\}$ in Eq.\ref{eq7} works as penalty terms in the Lagrange function (with Lagrangian multiplier $\alpha$) .}.

\begin{algorithm}[h]
   \caption{The CPED algorithm}
   \label{alg:example}
\begin{algorithmic}
   \STATE {\bfseries Input:} dataset $\cD$, target network update rate $\kappa$, Lagrange multiplier $\alpha$, training ratio of the generator and discrimination $r$, mini-batch size $N$, hyperparameters $\lambda, c$.
   \STATE Initialize generator $G_{\theta}$, discriminator $D_{\phi}$, loss function in GAN $\mathcal{L_G}(\cdot)$, behavior policy likelihood $L_\theta^{\pi_\beta}(\cdot)$, Q networks $\{Q_{\eta_1},Q_{\eta_2}\}$, actor $\pi_{\psi}$,  target networks $\{Q_{\eta_1'},Q_{\eta_2'}\}$, target actor $\pi_{\psi'}$, with $\psi'\leftarrow \psi, \eta_{i}'\leftarrow \eta_{i}, i=1,2$.
   \FOR{$i=1$ {\bfseries to} $M$}
   \STATE Sample mini-batch of transitions $(s, a, r, s') \sim \cD$
   \STATE \textbf{Initial Training Flow-GAN:}
   \STATE $\phi \leftarrow \argmax_{\phi}\mathcal{L_G}(G_{\theta},D_{\phi})$
   \STATE $\theta \leftarrow \argmin_{\theta}\max_{\phi}\mathcal{L_G}(G_{\theta},D_{\phi})-[\lambda\mathbb{E}_{(s,a) \sim \cD}\log(L_{\theta}^{\pi_{\beta}}(s,a))]$
    \ENDFOR

   \FOR{$i=1$ {\bfseries to} $N$}
   \STATE Sample mini-batch of transitions $(s, a, r, s') \sim \cD$
   \STATE \textbf{Updating Flow-GAN:}
   \STATE $\phi \leftarrow \argmax_{\phi}\mathcal{L_G}(G_{\theta},D_{\phi})$
   \STATE $\theta \leftarrow \argmin_{\theta}\max_{\phi}\mathcal{L_G}(G_{\theta},D_{\phi})-[\lambda\mathbb{E}_{(s,a) \sim \cD}\log(L_{\theta}^{\pi_{\beta}}(s,a))]$
   \STATE \textbf{Updating Q-function:}
   \STATE Get action for the next state, $\{\tilde{a} \sim \pi_{\psi'}(\cdot|s')+\varepsilon ,\varepsilon \sim clip(\mathcal{N}(0,\sigma),-c,c) \}$ 
   \STATE Let $y(s,a):=\min(Q_{\eta_{1}'}(s',\tilde{a}),Q_{\eta_{2}'}(s',\tilde{a}))$
   \STATE $\eta_{i} \leftarrow \argmin_{\eta_{i}}(Q_{\eta_{i}}(s,a)-(r+\gamma y(s,a)))^2, i=1,2$
   \STATE \textbf{Updating Actor:}
   \STATE Update $\psi$ according to Eq.\ref{eq7}, by using dual gradient descent with Lagrange multiplier $\alpha$
   \STATE \textbf{Update Target Networks:}
   \STATE $\psi'\leftarrow \kappa \psi+(1-\kappa)\psi';
   \eta_{i}'\leftarrow \kappa \eta_{i}+(1-\kappa) \eta_{i}', i=1,2$
    \ENDFOR
\end{algorithmic}
\end{algorithm}

The critic training is following the traditional Bellman equation as shown in Eq.\ref{eq_bellmanR}. The entire algorithm of CPED is summarized in Algorithm \ref{alg:example}.
In CPED, the Flow-GAN model plays a crucial role in estimating the behavior density. The density function is obtained from the generator model and is iteratively updated along with the discriminator. We further use the TD3 framework to solve the RL problem. It is noted that during Actor updates, the density of sampled actions should be higher than the threshold to avoid out-of-distribution areas. 

\section{Theoretical Analysis}

In this section, we provide theoretical analysis of our proposed method. In Section \ref{subsec_convergenceGAN}, we prove the convergence of GAN with the hybrid loss, and based on this result, we show the convergence of CPED in Section \ref{subsec_convCPED}.

\subsection{Convergence of GAN with the Hybrid Loss}\label{subsec_convergenceGAN}
In this subsection, we show that training GAN with the hybrid loss function Eq.\ref{eq4} can yield an accurate density estimator. For ease of presentation, assume we observed $X_j\sim p^* = \frac{{\rm d}\mu^*}{{\rm d} \nu}$, $j=1,...,n$, where $\mu^*$ is the underlying true probability measure (the probability measure of behavior policy), and $\nu$ is the Lebesgue measure. Motivated by \cite{Arjovsky17,liang2018generative,liu2017approximation}, we consider the Integral Probability Metric (IPM) defined as
\begin{equation}
        \label{eq:GAN_formulation}
         d_{\cF_d}(\mu_1,\mu_2):= \sup_{f\in \cF_d} \EE_{X\sim \mu_1} f(X)-\EE_{Y\sim \mu_2}f(Y),
\end{equation}
where $\mu_1$ and $\mu_2$ are two probability measures, and $\cF_d$ is a discriminator class. Under IPM, the GAN framework with the hybrid loss Eq.\ref{eq4} solves the empirical problem (cf., Eq. (1.1) of \cite{liang2018generative} for the original GAN loss function)
\begin{align}\label{eq_GAN}
     \mu_n \in \argmin_{\mu\in \cQ_g}\max_{f\in \cF_d} \int_\Omega f{\rm d}\mu - \int_\Omega f{\rm d}\tilde \mu_n - \lambda \int_\Omega \log p{\rm d}\hat \mu_n,
\end{align}
where $\hat \mu_n = \frac{1}{n}\sum_{j=1}^n \delta_{X_j}$ is the empirical measure, $\tilde \mu_n$ is a (regularized) density estimation, $\cQ_g$ is the generator class, and $p = \frac{{\rm d}\mu}{{\rm d} \nu}$. In the following (informal) theorem, we show that $\mu_n$ is an accurate estimator of the underlying true distribution. A detailed version of Theorem \ref{thm1} and its proof can be found in Appendix \ref{app_pfgan}.

\begin{thm}[Informal]\label{thm1}
Suppose the generator class $\cQ_g$ and the discriminator class $\cF_d$ are induced by some Sobolev spaces. Choose $\lambda$ as a constant. Under certain conditions, 
    \begin{align*}
    d_{\cF_d}(\mu^*,\mu_n) = O_{\PP}(n^{-1/2}), {\rm KL}(\mu^*||\mu_n) = O_{\PP}(n^{-1/2}),
\end{align*}
where $\mu^*$ is the true probability measure, $\mu_n$ is as in Eq.\ref{eq_GAN} with hybrid loss, and ${\rm KL}(\mu^*||\mu_n)$ is the KL divergence.
\end{thm}

Theorem \ref{thm1} states a fast convergence rate $O_{\PP}(n^{-1/2})$ of the IPM and KL divergence can be achieved. Therefore, applying the hybrid loss can provide not only a good generator, but also provide an accurate density estimate, which can be utilized in our CPED framework to control the safe region.

\begin{remark}
    Given the universal approximation theorem \cite{hornik1989multilayer,csaji2001approximation}, the neural networks can approximate any smooth function. Thus, we consider Sobolev function classes instead of neural networks for the ease of mathematical treatment. If both the generator and discriminator classes are neural networks, one may adopt similar approach as in \cite{liang2018generative,ding2020high} and then apply Theorem \ref{thm1}.
\end{remark}

\begin{remark}
    In \cite{liang2018generative}, the convergence rate of $\EE d_{\cF_d}(\mu^*,\mu_n)$ is considered, while we provide a sharper characterization of $d_{\cF_d}(\mu^*,\mu_n)$, by showing the convergence is in probability with certain convergence rate.
\end{remark}

\subsection{Convergence of CPED}\label{subsec_convCPED}

With an accurate estimate of the underlying density, in this subsection, we give a theoretical analysis of how CPED can find the optimal $Q$-function value over the product space of $\mathscr{S}\times \mathscr{A}$. The proof of Theorem \ref{thm2} can be found in Appendix \ref{appc}.

\newtheorem{thm2}{\bf Theorem}[section]
\begin{thm}\label{thm2}
Suppose $Q^*(s,a)$ is the optimal $Q$-value on product space  $\mathscr{S}\times \mathscr{A}$. Let $\hat{\pi}(\cdot|s):=\argmax_{a}Q^*(s,a)$ be a learning policy. Let $\hat{\pi}^{\Delta}(\cdot|s)$ be the probability-controlled learning policy using CPED, which is defined on the support of $\pi_{\beta}$. Suppose $\hat{\pi}^{\Delta}(\cdot|s)>0$ on $\mathscr{A}$.  Then with probability tending to one,
\begin{equation}\label{eq8}
Q^*(s,a)=Q(s,\hat{\pi}^{\Delta}(\cdot|s)).
\end{equation}
\end{thm}

Furthermore, when minimizing the Bellman residual to solve the optimal $Q$-function and training the policy by iteration, we have the following theorem, whose proof is provided in Appendix \ref{appd}.

\begin{thm}\label{thm3}
Suppose in step $k+1$, the learned policy is defined as $\pi_{k+1}:=\argmax_{a}Q^{\pi_{k}}(s,a),\forall s \in \mathscr{S}$. Let $V^{\pi_k}$ be the value function related to $\pi_k$ in step k, defined as $V^{\pi_k}(s) = \EE_{a\sim \pi_k(\cdot|s)} [Q(s,a)]$, and $V^*$ be the optimal value function over the product space $\mathscr{S}\times \mathscr{A}$. Let $\hat{\pi}^{\Delta}_{k}$ be the probability-controlled learning policy using CPED in step $k$, which is defined on the support of $\pi_{\beta}$. Then with probability tending to one,
\begin{equation}\label{eq9}
\Vert V^{\hat{\pi}^{\Delta}_{k+1}}-V^* \Vert_{\infty} \le \gamma^{k+1} \Vert V^{\hat{\pi}^{\Delta}_{0}}-V^* \Vert_{\infty}.
\end{equation}
\end{thm}

Theorem \ref{thm3} provides assurance that CPED can achieve the optimal value function with a linear convergence rate as the iteration number approaches infinity. Despite under offline settings, this rate is consistent with the standard rate in online RL \cite{Agarwal2019ReinforcementLT}. When combined with Theorems \ref{thm1} and \ref{thm2}, we can confidently conclude that CPED's effectiveness is theoretically guaranteed.

\section{Experiments}

In the experimental section, we analyze the performance of CPED on several standard offline RL tasks \cite{fu2020d4rl}. We compare CPED with a wide range of offline RL methods and analyze the performance of CPED. Further implementation details are provided in Appendix \ref{implemention}.

\subsection{Performance Comparison on Standard Benchmarking Datasets for Offline RL}

The proposed CPED algorithm is evaluated on the D4RL \cite{fu2020d4rl} Gym-MuJoCo and AntMaze tasks. Details of these tasks are shown in Appendix \ref{dataset}.
As a model-free method, CPED is compared with the following model-free baselines:  behavioral cloning (BC) method, BCQ \cite{Fujimoto2019}, DT \cite{chen2021decision}, AWAC \cite{nair2020awac}, BEAR
\cite{Kumar2019}, Onestep RL \cite{brandfonbrener2021offline},TD3+BC \cite{fujimoto2021minimalist}, CQL \cite{Kumar2020}, IQL \cite{kostrikov2021offline}, SPOT \cite{wu2022supported}, and PLAS \cite{zhou2021plas}. In the experiments, the performance of all alternatives are measured by the average normalized scores in an episode. The experimental results reported in this paper are either from the authors' original experiments or from our replications.

\begin{table*}[h!]
\caption{The performance of CPED and other competing methods on the various D4RL Gym-MuJoCo tasks. We report the performance of Average normalized scores and standard deviations over 5 random seeds. med = medium, r = replay, e = expert, hc = halfcheetah, wa = walker2d, ho=hopper  }
\label{tab1}
\vskip 0.15in
\begin{center}
\begin{small}
\begin{tabular}{l|lcccccccccr}
\toprule
Dataset& BC & AWAC & DT &  Onestep & TD3+BC & CQL & IQL & SPOT & CPED(Ours)\\
\midrule
hc-med    &42.6 &43.5 &42.6 &48.4 &48.3 &44.0 &47.4 &58.4& \pmb{61.8{\tiny $\pm$1.6}}  \\
ho-med     &52.9 &57.0 &67.6 &59.6 &59.3 &58.5 &66.2&86.0& \pmb{100.1{\tiny$\pm$2.8}}\\
wa-med      &75.3 &72.4 &74.0 &81.8 &83.7 &72.5 &78.3 &86.4& \pmb{90.2{\tiny $\pm$1.7}} \\
hc-med-r     &36.6 &40.5 &36.6 &38.1 &44.6 &45.5 &44.2 &52.2& \pmb{55.8{\tiny $\pm$2.9}} \\
ho-med-r   &18.1 &37.2 &82.7 &97.5 &60.9 &95.0 &94.7&\pmb{100.2}& 98.1{\tiny $\pm$2.1} \\
wa-med-r   &26.0 &27.0 &66.6 &49.5 &81.8 &77.2 &73.8 &91.6& \pmb{91.9{\tiny $\pm$0.9}}\\
hc-med-e     &55.2 &42.8 &86.8 &\pmb{93.4} &90.7 &91.6 &86.7 &86.9& 85.4{\tiny $\pm$10.9} \\
ho-med-e   &52.5 &55.8 &\pmb{107.6} &103.3 &98.0 &105.4 &91.5 &99.3& 95.3{\tiny $\pm$13.5} \\
wa-med-e   &107.5 &74.5 &108.1 &113.0 &110.1 &108.8 &109.6 &112.0& \pmb{113.04{\tiny $\pm$1.4}}\\
\midrule
Total  &466.7 &450.7 &672.6 &684.6 &677.4 &698.5 &692.4 &773.0& \pmb{791.7{\tiny $\pm$37.8}} \\
\bottomrule
\end{tabular}
\end{small}
\end{center}
\vskip -0.1in
\end{table*}

\begin{table*}[h!]
\caption{The performance of CPED and other competing methods on the various AntMaze tasks. We report the performance of Average normalized scores and standard deviations over 5 random seeds. med = medium, d = diverse, p = play. }
\label{tab2}
\vskip 0.15in
\begin{center}
\begin{small}
\begin{tabular}{l|lccccccccccr}
\toprule
Dataset & BCQ&BEAR&BC&DT&TD3+BC&PLAS&CQL&IQL& SPOT & CPED(Ours)\\
\midrule
umaze   &78.9&73.0&49.2&54.2&73.0& 62.0 &82.6& 89.6& 93.5& \pmb{96.8{\tiny$\pm$2.6}}  \\
umaze-d     &55.0 &61.0 &41.8 &41.2 &47.0 &45.4 &10.2 &\pmb{65.6} &40.7& 55.6{\tiny$\pm$2.2}\\
med-p      &0.0 &0.0 &0.4 &0.0 &0.0 &31.4 &59.0 &76.4 &74.7& \pmb{85.1{\tiny$\pm$3.4}} \\
med-d     &0.0 &8.0 &0.2 &0.0 &0.2 &20.6 &46.6 &72.8 &\pmb{79.1}& 72.1{\tiny$\pm$2.9} \\
large-p   &6.7 &0.0 &0.0 &0.0 &0.0 &2.2 &16.4 &\pmb{42.0} &35.3& 34.9{\tiny$\pm$5.3} \\
large-d   &2.2 &0.0 &0.0 &0.0 &0.0 &3.0 &3.2 &\pmb{46.0} &36.3& 32.3{\tiny$\pm$7.4}\\
\midrule
Total  &142.8 &142.0 &91.6 &95.4 &120.2 &164.6 &218.0 &\pmb{392.4} &359.6& 376.8{\tiny$\pm$23.8} \\
\bottomrule
\end{tabular}
\end{small}
\end{center}
\vskip -0.1in
\end{table*}

\begin{figure*}[h!]
\vskip 0.2in
\begin{center}
\centerline{\includegraphics[width=\linewidth]{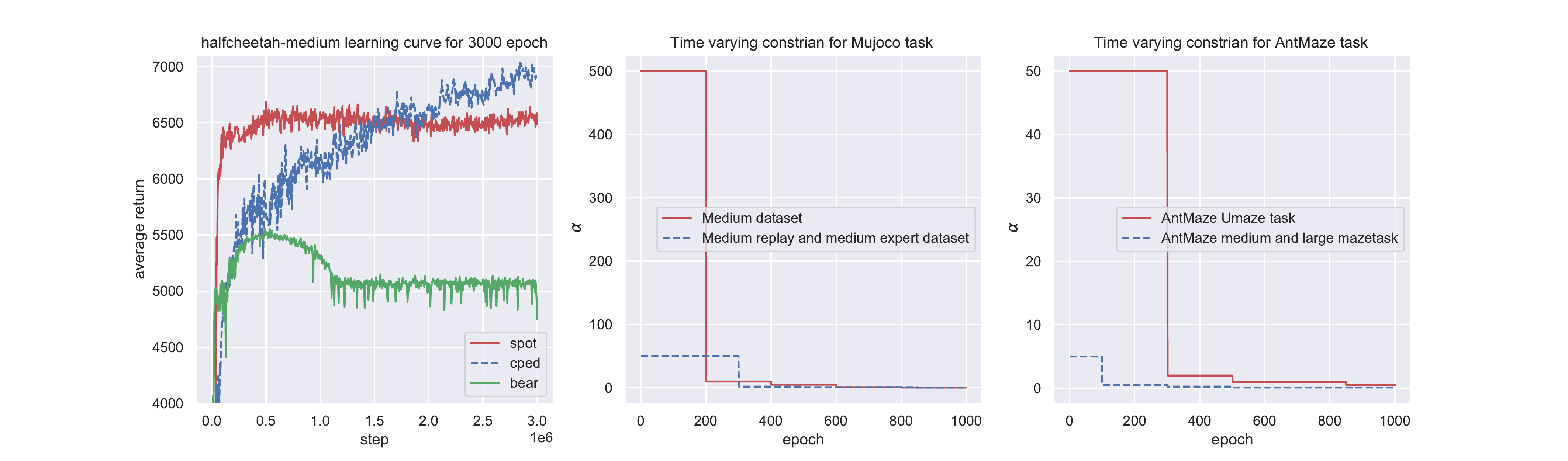}}

\caption{(a) Average performance of BEAR and CPED on halfcheetah-medium task averaged over 5 seeds. BEAR can reach a bottleneck very quickly. CPED remain increasing after reaching the bottleneck. (b) The time(epoch) varying constrain parameter $\alpha$ used in Gym-MuJoCo task. (c) The time(epoch) varying constrain parameter $\alpha$ used in AntMaze task}
\label{fig3}
\end{center}
\vskip -0.2in
\end{figure*}

Tables \ref{tab1} and \ref{tab2} show the experimental results of Gym-MuJoCo and AntMaze tasks, respectively. In Gym-MuJoCo tasks, CPED outperforms the competitors in most cases. In particular, our method has achieved remarkable success on the halfcheetah and walker2d tasks, especially under the medium-quality dataset. For the medium-expert task, CPED performs slightly inferior to other tasks, which is due to the fact that the expert task needs tight control so that the learning policy stays close to the behavior policy. In the AntMaze tasks, CPED performs slightly worse than IQL but outperforms the other baselines, and it achieves better returns under ``unmaze'' and ``medium-play'' tasks.

To further discuss the learning dynamics of the proposed methods, the learning curve of BEAR, SPOT, and CPED in halfcheetah-medium task is shown in Figure \ref{fig3}(a). It is noted that the average return of BEAR increases very fast and remains unchanged after a smooth decrease in subsequent training, indicating that policy exploration is limited.  This is consistent with the conservative nature of policy control methods that prevent the learning policy from straying too far from behavior policy. The pattern is similar to SPOT's learning curve, which shows a rapid increase and then enters a plateau. In contrast, the average returns of CPED methods continue to rise after a sharp increase to a certain extent, indicating that CPED keeps exploring the data and benefits from exploration. This also confirms Theorem \ref{thm3} (see Section \ref{appc}) from the experimental perspective.

\subsection{Ablation Study}
\label{ablation}

\textbf{Time varying hyperparameter $\alpha$ in policy control methods.} When improving actor using Eq.\ref{eq7}, a Lagrange hyperparameter $\alpha$ is needed for this constrained optimization problem. The larger the $\alpha$, the more restricted actions taken by the learning policy. Previous methods either use an auto-learning $\alpha$ \cite{Kumar2019,wu2021uncertainty} or a constant $\alpha$ \cite{wu2019behavior,wu2022supported}.  BRAC \cite{wu2019behavior} find that auto-learning $\alpha$ performs worse than the constant $\alpha$. In our experiment, we
apply a time (epoch) varying $\alpha$ scheme with piece-wise constant functions and find this scheme performs better than the constant setting. The rationale behind this time-varying scheme is that we employ strong constraints for the action taken at the early stage of policy training in case the model fails \cite{levine2020offline}. The constraint is further relaxed by decreasing $\alpha$ when the policy starts to converge. The piece-wise function of $\alpha$ for the Mujoco tasks and AntMaze tasks are plot in Figure \ref{fig3}(b)-(c), and the performance results of the Mujoco tasks are shown in Figure \ref{fig9} in in Appendix \ref{ablationinapp}. We can see that the Mujoco task significantly benefits from this time varying scheme and delivers better returns in most scenarios.

\textbf{Likelihood threshold $\epsilon$.} The hyperparameter $\epsilon$ determines the tightness of our estimated safety boundaries. A larger $\epsilon$ leads to more relaxed boundaries and versa vice. A commonly used setting for $\epsilon$ is the quantile of behavior policy density. However, the quantile value is difficult to set when the dimension is large. In our experiment, CPED uses the mean likelihood of the training batch as $\epsilon$, which decreases the effect of the estimation error of the behavior policy. See Appendix \ref{ablationinapp} for additional discussion and details.

\section{Conclusion and Future Work}

In this article, we present CPED, a novel approach for policy control in offline RL. CPED leverages the Flow-GAN model to estimate the density of the behavior policy, enabling the identification of safe areas and avoiding the learning of policies that visit OOD points. We validate the effectiveness of CPED both theoretically and experimentally. Theoretical analysis demonstrates that CPED has a high probability of achieving the optimal policy and generating large returns. Experimental results on the Gym-MuJoCo and AntMaze tasks demonstrate that CPED surpasses state-of-the-art competitors, highlighting its superior performance.

While CPED shows great capacity in experiment studies, there are several potential directions for its extension.. Firstly, as a generic method, GAN has diverse variants and it is necessary to try out more powerful GAN structures for density estimation. Secondly, when the feasible region is identified, how to explore the region efficiently remains another important issue, and we leave it to future work. Finally, it is an interesting direction to examine the performance of CPED in more complex scenarios, including dataset  generated by multiple behavior policies or multiple agents, etc.

\section*{Acknowledgements}
We would like to thank AC and reviewers for their valuable comments on the manuscript.
Wenjia Wang was supported by the Guangzhou-HKUST(GZ) Joint Funding Program (No. 2023A03J0019) and Guangzhou Municipal Science and Technology Project (No. 2023A03J0003)

\bibliography{references}

\begin{thebibliography}{}

\bibitem[Adams and Fournier, 2003]{adams2003sobolev}
Adams, R.~A. and Fournier, J.~J. (2003).
\newblock {\em Sobolev Spaces}, volume 140.
\newblock Academic press.

\bibitem[Afsar et~al., 2022]{Afsar2022}
Afsar, M.~M., Crump, T., and Far, B. (2022).
\newblock Reinforcement learning based recommender systems: A survey.
\newblock {\em ACM Computing Surveys}, 55(7):1--38.

\bibitem[Agarwal et~al., 2019]{Agarwal2019ReinforcementLT}
Agarwal, A., Jiang, N., and Kakade, S.~M. (2019).
\newblock Reinforcement learning: Theory and algorithms.

\bibitem[Agarwal et~al., 2020]{Agarwal2020}
Agarwal, R., Schuurmans, D., and Norouzi, M. (2020).
\newblock An optimistic perspective on offline reinforcement learning.
\newblock volume PartF168147-1.

\bibitem[An et~al., 2021]{an2021uncertainty}
An, G., Moon, S., Kim, J.-H., and Song, H.~O. (2021).
\newblock Uncertainty-based offline reinforcement learning with diversified
  q-ensemble.
\newblock {\em Advances in neural information processing systems},
  34:7436--7447.

\bibitem[Arjovsky et~al., 2017]{Arjovsky17}
Arjovsky, M., Chintala, S., and Bottou, L. (2017).
\newblock Wasserstein generative adversarial networks.
\newblock In {\em International Conference on Machine Learning}, pages
  214--223. PMLR.

\bibitem[Brandfonbrener et~al., 2021]{brandfonbrener2021offline}
Brandfonbrener, D., Whitney, W., Ranganath, R., and Bruna, J. (2021).
\newblock Offline rl without off-policy evaluation.
\newblock {\em Advances in neural information processing systems},
  34:4933--4946.

\bibitem[Chen et~al., 2022a]{chen2022offline}
Chen, H., Lu, C., Ying, C., Su, H., and Zhu, J. (2022a).
\newblock Offline reinforcement learning via high-fidelity generative behavior
  modeling.
\newblock {\em arXiv preprint arXiv:2209.14548}.

\bibitem[Chen et~al., 2021]{chen2021decision}
Chen, L., Lu, K., Rajeswaran, A., Lee, K., Grover, A., Laskin, M., Abbeel, P.,
  Srinivas, A., and Mordatch, I. (2021).
\newblock Decision transformer: Reinforcement learning via sequence modeling.
\newblock {\em Advances in neural information processing systems},
  34:15084--15097.

\bibitem[Chen et~al., 2022b]{chen2022latent}
Chen, X., Ghadirzadeh, A., Yu, T., Gao, Y., Wang, J., Li, W., Liang, B., Finn,
  C., and Zhang, C. (2022b).
\newblock Latent-variable advantage-weighted policy optimization for offline
  rl.
\newblock {\em arXiv preprint arXiv:2203.08949}.

\bibitem[Cs{\'a}ji et~al., 2001]{csaji2001approximation}
Cs{\'a}ji, B.~C. et~al. (2001).
\newblock Approximation with artificial neural networks.
\newblock {\em Faculty of Sciences, Etvs Lornd University, Hungary}, 24(48):7.

\bibitem[Degrave et~al., 2022]{Degrave2022}
Degrave, J., Felici, F., Buchli, J., Neunert, M., Tracey, B., Carpanese, F.,
  Ewalds, T., Hafner, R., Abdolmaleki, A., de~las Casas, D., Donner, C., Fritz,
  L., Galperti, C., Huber, A., Keeling, J., Tsimpoukelli, M., Kay, J., Merle,
  A., Moret, J.~M., Noury, S., Pesamosca, F., Pfau, D., Sauter, O., Sommariva,
  C., Coda, S., Duval, B., Fasoli, A., Kohli, P., Kavukcuoglu, K., Hassabis,
  D., and Riedmiller, M. (2022).
\newblock Magnetic control of tokamak plasmas through deep reinforcement
  learning.
\newblock {\em Nature}, 602.

\bibitem[Ding et~al., 2020]{ding2020high}
Ding, L., Zou, L., Wang, W., Shahrampour, S., and Tuo, R. (2020).
\newblock High-dimensional non-parametric density estimation in mixed smooth
  sobolev spaces.
\newblock {\em arXiv preprint arXiv:2006.03696}.

\bibitem[Dinh et~al., 2015]{Dinh2015}
Dinh, L., Krueger, D., and Bengio, Y. (2015).
\newblock Nice: Non-linear independent components estimation.

\bibitem[Dinh et~al., 2017]{Dinh2017}
Dinh, L., Sohl-Dickstein, J., and Bengio, S. (2017).
\newblock Density estimation using real nvp.

\bibitem[Doan~T, 2018]{Doan2018ganqlearning}
Doan~T, M.~B. (2018).
\newblock Gan q-learning.
\newblock {\em arXiv preprint arXiv:1805.04874}.

\bibitem[Donoho et~al., 1995]{donoho1995wavelet}
Donoho, D.~L., Johnstone, I.~M., Kerkyacharian, G., and Picard, D. (1995).
\newblock Wavelet shrinkage: asymptopia?
\newblock {\em Journal of the Royal Statistical Society: Series B
  (Methodological)}, 57(2):301--337.

\bibitem[Donoho et~al., 1996]{Donoho96}
Donoho, D.~L., Johnstone, I.~M., Kerkyacharian, G., and Picard, D. (1996).
\newblock Density estimation by wavelet thresholding.
\newblock {\em The Annals of Statistics}, 24(2):508--539.

\bibitem[Finn et~al., 2016a]{finn2016connection}
Finn, C., Christiano, P., Abbeel, P., and Levine, S. (2016a).
\newblock A connection between generative adversarial networks, inverse
  reinforcement learning, and energy-based models.
\newblock {\em arXiv preprint arXiv:1611.03852}.

\bibitem[Finn et~al., 2016b]{Finn2016}
Finn, C., Levine, S., and Abbeel, P. (2016b).
\newblock Guided cost learning: Deep inverse optimal control via policy
  optimization.
\newblock volume~1.

\bibitem[Friedman et~al., 2001]{friedman2001elements}
Friedman, J., Hastie, T., and Tibshirani, R. (2001).
\newblock {\em The Elements of Statistical Learning}, volume~1.
\newblock Springer series in statistics New York, NY, USA:.

\bibitem[Fu et~al., 2020]{fu2020d4rl}
Fu, J., Kumar, A., Nachum, O., Tucker, G., and Levine, S. (2020).
\newblock D4rl: Datasets for deep data-driven reinforcement learning.
\newblock {\em arXiv preprint arXiv:2004.07219}.

\bibitem[Fujimoto and Gu, 2021]{fujimoto2021minimalist}
Fujimoto, S. and Gu, S.~S. (2021).
\newblock A minimalist approach to offline reinforcement learning.
\newblock {\em Advances in neural information processing systems},
  34:20132--20145.

\bibitem[Fujimoto et~al., 2018]{Fujimoto2018}
Fujimoto, S., Hoof, H.~V., and Meger, D. (2018).
\newblock Addressing function approximation error in actor-critic methods.
\newblock volume~4.

\bibitem[Fujimoto et~al., 2019]{Fujimoto2019}
Fujimoto, S., Meger, D., and Precup, D. (2019).
\newblock Off-policy deep reinforcement learning without exploration.
\newblock volume 2019-June.

\bibitem[Garg et~al., 2023]{garg2023extreme}
Garg, D., Hejna, J., Geist, M., and Ermon, S. (2023).
\newblock Extreme q-learning: Maxent rl without entropy.
\newblock {\em arXiv preprint arXiv:2301.02328}.

\bibitem[Ghasemipour et~al., 2021]{ghasemipour2021emaq}
Ghasemipour, S. K.~S., Schuurmans, D., and Gu, S.~S. (2021).
\newblock Emaq: Expected-max q-learning operator for simple yet effective
  offline and online rl.
\newblock In {\em International Conference on Machine Learning}, pages
  3682--3691. PMLR.

\bibitem[Goodfellow et~al., 2014]{Goodfellow14}
Goodfellow, I., Pouget-Abadie, J., Mirza, M., Xu, B., Warde-Farley, D., Ozair,
  S., Courville, A., and Bengio, Y. (2014).
\newblock Generative adversarial nets.
\newblock {\em Advances in Neural Information Processing Systems}, 27.

\bibitem[Grover et~al., 2018]{Grover2018}
Grover, A., Dhar, M., and Ermon, S. (2018).
\newblock Flow-gan: Combining maximum likelihood and adversarial learning in
  generative models.

\bibitem[Gulrajani et~al., 2017]{gulrajani2017improved}
Gulrajani, I., Ahmed, F., Arjovsky, M., Dumoulin, V., and Courville, A.~C.
  (2017).
\newblock Improved training of wasserstein gans.
\newblock {\em Advances in neural information processing systems}, 30.

\bibitem[Haarnoja et~al., 2018]{Haarnoja2018}
Haarnoja, T., Zhou, A., Abbeel, P., and Levine, S. (2018).
\newblock Soft actor-critic: Off-policy maximum entropy deep reinforcement
  learning with a stochastic actor.
\newblock volume~5.

\bibitem[Hall, 1981]{Hall1981}
Hall, P. (1981).
\newblock On trigonometric series estimates of densities.
\newblock {\em The Anuals of Statistics}, 9(3):683–685.

\bibitem[Hansen-Estruch et~al., 2023]{hansen2023idql}
Hansen-Estruch, P., Kostrikov, I., Janner, M., Kuba, J.~G., and Levine, S.
  (2023).
\newblock Idql: Implicit q-learning as an actor-critic method with diffusion
  policies.
\newblock {\em arXiv preprint arXiv:2304.10573}.

\bibitem[Hathaway, 1985]{hathaway1985}
Hathaway, R. (1985).
\newblock A constrained formulation of maximum-likelihood estimation for normal
  mixture distributions.
\newblock {\em The Annals of Statistics}, 13(2):795--800.

\bibitem[Ho and Ermon, 2016]{ho2016generative}
Ho, J. and Ermon, S. (2016).
\newblock Generative adversarial imitation learning.
\newblock {\em Advances in neural information processing systems}, 29.

\bibitem[Ho et~al., 2016]{Ho2016}
Ho, J., Gupta, J.~K., and Ermon, S. (2016).
\newblock Model-free imitation learning with policy optimization.
\newblock volume~6.

\bibitem[Ho et~al., 2020]{ho2020denoising}
Ho, J., Jain, A., and Abbeel, P. (2020).
\newblock Denoising diffusion probabilistic models.
\newblock {\em Advances in Neural Information Processing Systems},
  33:6840--6851.

\bibitem[Hornik et~al., 1989]{hornik1989multilayer}
Hornik, K., Stinchcombe, M., and White, H. (1989).
\newblock Multilayer feedforward networks are universal approximators.
\newblock {\em Neural networks}, 2(5):359--366.

\bibitem[Janner et~al., 2022]{janner2022planning}
Janner, M., Du, Y., Tenenbaum, J.~B., and Levine, S. (2022).
\newblock Planning with diffusion for flexible behavior synthesis.
\newblock {\em arXiv preprint arXiv:2205.09991}.

\bibitem[Jaques et~al., 2019]{jaques2019way}
Jaques, N., Ghandeharioun, A., Shen, J.~H., Ferguson, C., Lapedriza, A., Jones,
  N., Gu, S., and Picard, R. (2019).
\newblock Way off-policy batch deep reinforcement learning of implicit human
  preferences in dialog.
\newblock {\em arXiv preprint arXiv:1907.00456}.

\bibitem[Johnson et~al., 2016]{Johnson2016}
Johnson, A.~E., Pollard, T.~J., Shen, L., Lehman, L. W.~H., Feng, M., Ghassemi,
  M., Moody, B., Szolovits, P., Celi, L.~A., and Mark, R.~G. (2016).
\newblock Mimic-iii, a freely accessible critical care database.
\newblock {\em Scientific Data}, 3.

\bibitem[Kalashnikov et~al., 2018]{kalashnikov2018scalable}
Kalashnikov, D., Irpan, A., Pastor, P., Ibarz, J., Herzog, A., Jang, E.,
  Quillen, D., Holly, E., Kalakrishnan, M., Vanhoucke, V., et~al. (2018).
\newblock Scalable deep reinforcement learning for vision-based robotic
  manipulation.
\newblock In {\em Conference on Robot Learning}, pages 651--673. PMLR.

\bibitem[Kostrikov et~al., 2021]{kostrikov2021offline}
Kostrikov, I., Nair, A., and Levine, S. (2021).
\newblock Offline reinforcement learning with implicit q-learning.
\newblock {\em arXiv preprint arXiv:2110.06169}.

\bibitem[Kumar et~al., 2019]{Kumar2019}
Kumar, A., Fu, J., Tucker, G., and Levine, S. (2019).
\newblock Stabilizing off-policy q-learning via bootstrapping error reduction.
\newblock volume~32.

\bibitem[Kumar et~al., 2020]{Kumar2020}
Kumar, A., Zhou, A., Tucker, G., and Levine, S. (2020).
\newblock Conservative q-learning for offline reinforcement learning.
\newblock volume 2020-December.

\bibitem[Kuznetsov et~al., 2020]{kuznetsov2020controlling}
Kuznetsov, A., Shvechikov, P., Grishin, A., and Vetrov, D. (2020).
\newblock Controlling overestimation bias with truncated mixture of continuous
  distributional quantile critics.
\newblock In {\em International Conference on Machine Learning}, pages
  5556--5566. PMLR.

\bibitem[Levine et~al., 2016]{levine2016end}
Levine, S., Finn, C., Darrell, T., and Abbeel, P. (2016).
\newblock End-to-end training of deep visuomotor policies.
\newblock {\em The Journal of Machine Learning Research}, 17(1):1334--1373.

\bibitem[Levine et~al., 2020]{levine2020offline}
Levine, S., Kumar, A., Tucker, G., and Fu, J. (2020).
\newblock Offline reinforcement learning: Tutorial, review, and perspectives on
  open problems.
\newblock {\em arXiv preprint arXiv:2005.01643}.

\bibitem[Li et~al., 2017]{li2017mmd}
Li, C.-L., Chang, W.-C., Cheng, Y., Yang, Y., and P{\'o}czos, B. (2017).
\newblock {MMD GAN}: {T}owards deeper understanding of moment matching network.
\newblock In {\em Advances in Neural Information Processing Systems}, pages
  2203--2213.

\bibitem[Li et~al., 2022]{li2022doge}
Li, J., Zhan, X., Xu, H., Zhu, X., Liu, J., and Zhang, Y.-Q. (2022).
\newblock Distance-sensitive offline reinforcement learning.
\newblock {\em arXiv preprint arXiv:2205.11027}.

\bibitem[Li, 2019]{li2019rlapplication}
Li, Y. (2019).
\newblock Reinforcement learning applications.
\newblock {\em arXiv preprint arXiv:1908.06973}.

\bibitem[Liang, 2021]{liang2018generative}
Liang, T. (2021).
\newblock How well generative adversarial networks learn distributions.
\newblock {\em Journal of Machine Learning Research, to appear}.

\bibitem[Lillicrap et~al., 2015]{Lillicrap2015}
Lillicrap, T.~P., Hunt, J.~J., Pritzel, A., Heess, N., Erez, T., Tassa, Y.,
  Silver, D., and Wierstra, D. (2015).
\newblock Continuous control with deep reinforcement learning.
\newblock {\em arXiv preprint arXiv:1509.02971}.

\bibitem[Liu et~al., 2017]{liu2017approximation}
Liu, S., Bousquet, O., and Chaudhuri, K. (2017).
\newblock Approximation and convergence properties of generative adversarial
  learning.
\newblock {\em arXiv preprint arXiv:1705.08991}.

\bibitem[Lyu et~al., 2022]{lyu2022mildly}
Lyu, J., Ma, X., Li, X., and Lu, Z. (2022).
\newblock Mildly conservative q-learning for offline reinforcement learning.
\newblock {\em arXiv preprint arXiv:2206.04745}.

\bibitem[Meyer, 1992]{meyer1992wavelets}
Meyer, Y. (1992).
\newblock {\em Wavelets and Operators: Volume 1}.
\newblock Number~37. Cambridge university press.

\bibitem[Mnih et~al., 2013]{Mnih2013}
Mnih, V., Kavukcuoglu, K., Silver, D., Graves, A., Antonoglou, I., Wierstra,
  D., and Riedmiller, M. (2013).
\newblock Playing atari with deep reinforcement learning.
\newblock {\em arXiv preprint arXiv:1312.5602}.

\bibitem[Nair et~al., 2020]{nair2020awac}
Nair, A., Gupta, A., Dalal, M., and Levine, S. (2020).
\newblock Awac: Accelerating online reinforcement learning with offline
  datasets.
\newblock {\em arXiv preprint arXiv:2006.09359}.

\bibitem[Ng et~al., 2000]{ng2000algorithms}
Ng, A.~Y., Russell, S., et~al. (2000).
\newblock Algorithms for inverse reinforcement learning.
\newblock In {\em Icml}, volume~1, page~2.

\bibitem[Radford et~al., 2015]{radford2015unsupervised}
Radford, A., Metz, L., and Chintala, S. (2015).
\newblock Unsupervised representation learning with deep convolutional
  generative adversarial networks.
\newblock {\em arXiv preprint arXiv:1511.06434}.

\bibitem[Scherrer et~al., 2015]{Scherrer2015}
Scherrer, B., Ghavamzadeh, M., Gabillon, V., Lesner, B., and Geist, M. (2015).
\newblock Approximate modified policy iteration and its application to the game
  of tetris.
\newblock {\em Journal of Machine Learning Research}, 16.

\bibitem[Schulman et~al., 2017]{Schulman2017}
Schulman, J., Wolski, F., Dhariwal, P., Radford, A., and Klimov, O. (2017).
\newblock Proximal policy optimization algorithms.
\newblock {\em arXiv preprint arXiv:1707.06347}.

\bibitem[Silver et~al., 2014]{Silver2014}
Silver, D., Lever, G., Heess, N., Degris, T., Wierstra, D., and Riedmiller, M.
  (2014).
\newblock Deterministic policy gradient algorithms.
\newblock In {\em International conference on machine learning}, pages
  387--395. PMLR.

\bibitem[Singh et~al., 2020]{singh2020parrot}
Singh, A., Liu, H., Zhou, G., Yu, A., Rhinehart, N., and Levine, S. (2020).
\newblock Parrot: Data-driven behavioral priors for reinforcement learning.
\newblock {\em arXiv preprint arXiv:2011.10024}.

\bibitem[Sutton and Barto, 1998]{Sutton1998}
Sutton, R. and Barto, A. (1998).
\newblock Reinforcement learning: An introduction.
\newblock {\em IEEE Transactions on Neural Networks}, 9.

\bibitem[Sutton and Barto, 2018]{sutton2018rlintro}
Sutton, R.~S. and Barto, A.~G. (2018).
\newblock {\em Reinforcement learning: An introduction}.
\newblock MIT press.

\bibitem[Urp{\'\i} et~al., 2021]{urpi2021risk}
Urp{\'\i}, N.~A., Curi, S., and Krause, A. (2021).
\newblock Risk-averse offline reinforcement learning.
\newblock {\em arXiv preprint arXiv:2102.05371}.

\bibitem[van~de Geer, 2000]{geer2000empirical}
van~de Geer, S. (2000).
\newblock {\em Empirical Processes in M-estimation}.
\newblock Cambridge University Press.

\bibitem[Van~Hasselt et~al., 2016]{Hasselt2016}
Van~Hasselt, H., Guez, A., and Silver, D. (2016).
\newblock Deep reinforcement learning with double q-learning.
\newblock In {\em Proceedings of the AAAI conference on artificial
  intelligence}, volume~30.

\bibitem[Wahba, 1981]{wahba1981}
Wahba, G. (1981).
\newblock Data-based optimal smoothing of orthogonal sereis density estimates.
\newblock {\em The Annals of Statistics}, 9(1):146--156.

\bibitem[Wang and Wang, 2015]{wang2015}
Wang, X. and Wang, Y. (2015).
\newblock Nonparametric multivariate density estimation using mixtures.
\newblock {\em Statistics and Computing}, 25(2):349--364.

\bibitem[Wang et~al., 2022]{wang2022diffusion}
Wang, Z., Hunt, J.~J., and Zhou, M. (2022).
\newblock Diffusion policies as an expressive policy class for offline
  reinforcement learning.
\newblock {\em arXiv preprint arXiv:2208.06193}.

\bibitem[Wang et~al., 2016]{wang2016}
Wang, Z., Schaul, T., Hessel, M., Hasselt, H., Lanctot, M., and Freitas, N.
  (2016).
\newblock Dueling network architectures for deep reinforcement learning.
\newblock In {\em International conference on machine learning}, pages
  1995--2003. PMLR.

\bibitem[White, 1982]{white1982maximum}
White, H. (1982).
\newblock Maximum likelihood estimation of misspecified models.
\newblock {\em Econometrica: Journal of the econometric society}, pages 1--25.

\bibitem[Wu et~al., 2022]{wu2022supported}
Wu, J., Wu, H., Qiu, Z., Wang, J., and Long, M. (2022).
\newblock Supported policy optimization for offline reinforcement learning.
\newblock {\em arXiv preprint arXiv:2202.06239}.

\bibitem[Wu et~al., 2019]{wu2019behavior}
Wu, Y., Tucker, G., and Nachum, O. (2019).
\newblock Behavior regularized offline reinforcement learning.
\newblock {\em arXiv preprint arXiv:1911.11361}.

\bibitem[Wu et~al., 2021]{wu2021uncertainty}
Wu, Y., Zhai, S., Srivastava, N., Susskind, J., Zhang, J., Salakhutdinov, R.,
  and Goh, H. (2021).
\newblock Uncertainty weighted actor-critic for offline reinforcement learning.
\newblock {\em arXiv preprint arXiv:2105.08140}.

\bibitem[Xiao et~al., 2023]{xiao2023sample}
Xiao, C., Wang, H., Pan, Y., White, A., and White, M. (2023).
\newblock The in-sample softmax for offline reinforcement learning.
\newblock {\em arXiv preprint arXiv:2302.14372}.

\bibitem[Xu et~al., 2023]{xu2023offline}
Xu, H., Jiang, L., Li, J., Yang, Z., Wang, Z., Chan, V. W.~K., and Zhan, X.
  (2023).
\newblock Offline rl with no ood actions: In-sample learning via implicit value
  regularization.
\newblock {\em arXiv preprint arXiv:2303.15810}.

\bibitem[Yang et~al., 2022a]{yang2022regularizing}
Yang, S., Feng, Y., Zhang, S., and Zhou, M. (2022a).
\newblock Regularizing a model-based policy stationary distribution to
  stabilize offline reinforcement learning.
\newblock In {\em International Conference on Machine Learning}, pages
  24980--25006. PMLR.

\bibitem[Yang et~al., 2022b]{yang2022behavior}
Yang, S., Wang, Z., Zheng, H., Feng, Y., and Zhou, M. (2022b).
\newblock A behavior regularized implicit policy for offline reinforcement
  learning.
\newblock {\em arXiv preprint arXiv:2202.09673}.

\bibitem[Yang et~al., 2022c]{yang2022regularized}
Yang, S., Wang, Z., Zheng, H., Feng, Y., and Zhou, M. (2022c).
\newblock A regularized implicit policy for offline reinforcement learning.
\newblock {\em arXiv preprint arXiv:2202.09673}.

\bibitem[Yarats et~al., 2021]{Yarats2021vae}
Yarats, D., Zhang, A., Kostrikov, I., Amos, B., Pineau, J., and Fergus, R.
  (2021).
\newblock Improving sample efficiency in model-free reinforcement learning from
  images.
\newblock In {\em Proceedings of the AAAI Conference on Artificial
  Intelligence}, volume~35, pages 10674--10681.

\bibitem[Yu et~al., 2018]{Yu2018}
Yu, F., Xian, W., Chen, Y., Liu, F., Liao, M., Madhavan, V., and Darrell, T.
  (2018).
\newblock Bdd100k: A diverse driving video database with scalable annotation
  tooling.
\newblock {\em Arxiv}.

\bibitem[Zhang et~al., 2022]{zhang2022sample}
Zhang, H., Mao, Y., Wang, B., He, S., Xu, Y., and Ji, X. (2022).
\newblock In-sample actor critic for offline reinforcement learning.
\newblock In {\em The Eleventh International Conference on Learning
  Representations}.

\bibitem[Zhou et~al., 2021]{zhou2021plas}
Zhou, W., Bajracharya, S., and Held, D. (2021).
\newblock Plas: Latent action space for offline reinforcement learning.
\newblock In {\em Conference on Robot Learning}, pages 1719--1735. PMLR.

\bibitem[Ziebart et~al., 2008]{ziebart2008maximum}
Ziebart, B.~D., Maas, A.~L., Bagnell, J.~A., Dey, A.~K., et~al. (2008).
\newblock Maximum entropy inverse reinforcement learning.
\newblock In {\em Aaai}, volume~8, pages 1433--1438. Chicago, IL, USA.

\end{thebibliography}
\bibliographystyle{apalike}

\appendix
\numberwithin{equation}{section}

\section*{Appendix}

\section{Convergence of GAN with the hybrid loss}\label{app_pfgan}

Before presenting the formal version of Theorem \ref{thm1} and its proof, we introduce some preliminaries. As stated in Theorem \ref{thm1}, we assume that both the discriminator class $\cF_d$ and the generator class $\cQ_g$ are within some Sobolev spaces. We define the Sobolev space via the wavelet bases as in nonparametric statistics \cite{donoho1995wavelet}.

Without loss of generality, let $\Omega=[0,1]^d$. Let $\phi$ be a ``farther wavelet'' satisfying $r$-regularity (see \cite{meyer1992wavelets} for details). For $j\in \NN$ and $\bs \in [2^j]^d$, where $[N]=\{1,...,N\}$, define 
\begin{align*}
\phi_{j,\bs}(\bx) =\left\{
    \begin{array}{ll}
        2^{dj}\phi(2^{dj}\bx - \bs), & \mbox{ if } 2^{dj}\bx - \bs\in \Omega,\\
        0, & {\rm otherwise}.
    \end{array}\right.
\end{align*}
Then, it can be shown that $\{\phi_{j,\bx}\}_{j\in \NN,\bs \in [2^j]^d}$ forms an orthonormal basis and for each $j\in \NN$, if $\bs\neq \bs'$, then $\phi_{j,\bs}(\bx)$ and $\phi_{j,\bs'}(\bx)$ have disjoint supports \cite{meyer1992wavelets}. A Sobolev space with smoothness $m$ can be defined as
\begin{align*}
    \cW^m(\Omega) =\{f\in L_2(\Omega):\|f\|_{\cW^m(\Omega)}<\infty\},
\end{align*}
where 
\begin{align*}
    \|f\|_{\cW^m(\Omega)}^2 = \sum_{j\in\NN}2^{jdm}\left(\sum_{\bs \in [2^j]^d} \langle f, \phi_{j,\bs}\rangle_{L_2(\Omega)}^2\right).
\end{align*}
We further define a ball with radius $R$ in $\cW^m(\Omega)$ as
\begin{align*}
    \cW^m(R) =\{f\in L_2(\Omega):\|f\|_{\cW^m(\Omega)}\leq R\}.
\end{align*}

Now we are ready to present a formal version of Theorem \ref{thm1} as follows. With an abuse of notation, we write $\mu \in \cQ_g$ to denote that the density function induced by $\mu$ is in $\cQ_g$. Thus, $\mu \in \cQ_g$ is the same as $p = \frac{{\rm d}\mu}{{\rm d}\nu} \in \cQ_g$, where $\nu$ is the Lebesgue measure. With this notation, we can directly write $\cQ_g$ and $\cF_d$ are subsets of some Sobolev spaces respectively instead of writing ``$\cQ_g$ and $\cF_d$ are induced by some Sobolev spaces'' as in the informal version Theorem \ref{thm1}, which simplifies the statement and proof.

\begin{thm}\label{thm_GAN}
Suppose the generator class $\cQ_g\subset \cW^{m_1}(R_1)$ and the discriminator class $\cF_d\subset \cW^{m_2}(R_2)$ are two Sobolev spaces with $m_1,m_2>d/2$, and both $\cQ_g$ and $\cF_d$ are symmetric. Suppose the underlying true density function $p^*\in \cQ_g$. Furthermore, assume that $\cG := \left\{g:g = \log (p/p^*) \right\}\subset \cW^{m_1}(R_3)$ for some constant $R_3$. Let $\lambda$ be a constant.
Then we have
\begin{align*}
    d_{\cF_d}(\mu^*, \mu_n) = O_{\PP}(n^{-1/2}), {\rm KL}(\mu^*||\mu_n) = O_{\PP}(n^{-1/2})
\end{align*}
where $\mu^*$ is the true probability measure, $\mu_n$ is as in Eq.\ref{eq_GAN} with hybrid loss, and ${\rm KL}(\mu^*||\mu_n)$ is the Kullback–Leibler divergence.
\end{thm}
\begin{remark}
    We assume $m_1,m_2>d/2$ because the Sobolev embedding theorem implies that all functions in the corresponding spaces are continuous.
\end{remark}

\textit{Proof of Theorem \ref{thm_GAN}.} 
Because $p^*\in \cQ_g$ and $\mu_n$ is as in Eq.\ref{eq_GAN}, we have
\begin{align}\label{eq_GANapp}
    d_{\cF_d}(\mu_n,\tilde \mu_n)- \lambda \int_\Omega \log p_n{\rm d}\hat \mu_n
    \leq  d_{\cF_d}(\mu^*,\tilde \mu_n) - \lambda \int_\Omega \log p^* {\rm d}\hat \mu_n,
\end{align}
where $p_n =\frac{{\rm d \mu_n}}{{\rm d}\nu}$ with $\nu$ the Lebesgue measure.
By the basic inequality $\sup_{\bx\in \Omega} (f(\bx) + g(\bx)) \leq \sup_{\bx\in \Omega} f(\bx) + \sup_{\bx\in \Omega} g(\bx)$, Eq.\ref{eq_GANapp} implies
\begin{align}\label{eq_GANapp2}
    d_{\cF_d}(\mu_n, \mu^*) - \lambda \int_\Omega \log p_n{\rm d}\hat \mu_n \leq & d_{\cF_d}(\mu^*, \tilde\mu_n) + d_{\cF_d}(\mu_n, \tilde\mu_n) - \lambda \int_\Omega \log p_n{\rm d}\hat \mu_n \nonumber\\
    \leq & 2 d_{\cF_d}(\mu^*,\tilde \mu_n)  - \lambda \int_\Omega \log p^* {\rm d}\hat \mu_n,
\end{align}
where we also use the assumption that both $\cQ_g$ and $\cF_d$ are symmetric in the first inequality.

By 
\begin{align*}
    d_{\cF_d}(\mu_n, \mu^*) \geq 0,
\end{align*}
Eq.\ref{eq_GANapp2} gives us
\begin{align}\label{eq_GANapp3}
    - \lambda \int_\Omega \log p_n(\bx){\rm d}\hat \mu_n \leq 2 d_{\cF_d}(\mu^*,\tilde \mu_n)  - \lambda \int_\Omega \log p^*(\bx) {\rm d}\hat \mu_n.
\end{align}

Since $\{\phi_{j,\bs}, j\in \NN, \bs \in [2^j]^d\}$ forms an orthogonal basis in $L_2(\Omega)$, for any functions  $f\in W^{m_2}(\Omega)$ and $p\in W^{m_1}(\Omega)$, they have the expansion as
\begin{align}\label{eq_ganpf_expansion}
    f(\bx) = & \sum_{j\in\NN}\sum_{\bs \in [2^j]^d} \langle f, \phi_{j,\bs}\rangle_{L_2(\Omega)}\phi_{j,\bs}(\bx),\nonumber\\
    p(\bx) = & \sum_{j\in\NN}\sum_{\bs \in [2^j]^d} \langle p, \phi_{j,\bs}\rangle_{L_2(\Omega)}\phi_{j,\bs}(\bx).
\end{align}
Let
\begin{align}\label{eq_ganpf_p1n}
    p_{1,n}(\bx) = \sum_{j=1}^M \sum_{\bs \in [2^j]^d} b_{j,\bs}\phi_{j,\bs}(\bx),
\end{align}
where $M$ will be determined later, and 
\begin{align*}
    b_{j,\bs} = \frac{1}{n}\sum_{k=1}^n \phi_{j,\bs}(X_k).
\end{align*} 
Thus, plugging Eq.\ref{eq_ganpf_expansion} and Eq.\ref{eq_ganpf_p1n} into $d_{\cF_d}(\mu^*,\tilde \mu_n)$ yields 
\begin{align}\label{eq_ganpf_dfd}
   & d_{\cF_d}(\mu^*,\tilde \mu_n)\nonumber\\
   =  & \sup_{f\in \cB_{W^{m_2}(\Omega)}(1)}\sum_{j=1}^M \sum_{\bs \in [2^j]^d} \langle f, \phi_{j,\bs}\rangle_{L_2(\Omega)}(b_{j,\bs} - \langle p, \phi_{j,\bs}\rangle_{L_2(\Omega)}) + \sum_{j=M+1}^\infty\sum_{\bs \in [2^j]^d} \langle f, \phi_{j,\bs}\rangle_{L_2(\Omega)}\langle p, \phi_{j,\bs}\rangle_{L_2(\Omega)}\nonumber\\
    = & I_1 + I_2.
\end{align}
The term $I_2$ is the truncation error, which can be bounded by
\begin{align}\label{eq_ganpf_I2}
    I_2 \leq & \left(\sum_{j=M+1}^\infty\sum_{\bs \in [2^j]^d} \langle f, \phi_{j,\bs}\rangle_{L_2(\Omega)}^2\right)^{1/2} \left(\sum_{j=M+1}^\infty\sum_{\bs \in [2^j]^d} \langle p, \phi_{j,\bs}\rangle_{L_2(\Omega)}\right)^{1/2}\nonumber\\
    \leq & 2^{-(Mdm_1+Mdm_2)/2}\left(\sum_{j=M+1}^\infty 2^{jdm_2}\sum_{\bs \in [2^j]^d} \langle f, \phi_{j,\bs}\rangle_{L_2(\Omega)}^2\right)^{1/2} \left(\sum_{j=M+1}^\infty 2^{jdm_1}\sum_{\bs \in [2^j]^d} \langle p, \phi_{j,\bs}\rangle_{L_2(\Omega)}\right)^{1/2}\nonumber\\
    \leq & C_1 2^{-(Mdm_1+Mdm_2)/2},
\end{align}
where the first inequality is by the Cauchy-Schwarz inequality, and the last inequality is because $f\in \cW^{m_2}(R_2)$ and $p\in \cW^{m_1}(R_1)$.

Next, we consider $I_1$. It can be checked that
\begin{align}\label{eq_ganpf_I1}
    I_1 = & \int_{\Omega} \sum_{j=1}^M \sum_{\bs \in [2^j]^d} \langle f, \phi_{j,\bs}\rangle_{L_2(\Omega)}\phi_{j,\bs}{\rm d} (\mu^* - \hat\mu_n).
\end{align}
Since $m_2>d/2$, we can see that the function $\|f_M\|_{L_\infty} < R$ for all $M>1$, where
\begin{align*}
    f_M := \sum_{j=1}^M \sum_{\bs \in [2^j]^d} \langle f, \phi_{j,\bs}\rangle_{L_2(\Omega)}\phi_{j,\bs},
\end{align*}
which, together with Lemma 5.11 of \cite{geer2000empirical}, implies that
\begin{align}\label{eq_ganpf_fmp}
    \left|\int_{\Omega} f_M {\rm d} (\mu^*-\hat \mu_n)\right| = O_{\PP}(n^{-1/2}),
\end{align}
where the right-hand-side term does not depend on $M$. Combining Eq.\ref{eq_ganpf_fmp} and Eq.\ref{eq_ganpf_I1}, the term $I_1$ can be bounded by
\begin{align}\label{eq_ganpf_I1X}
    I_1 = O_{\PP}(n^{-1/2}).
\end{align}
Plugging  Eq.\ref{eq_ganpf_I2} and Eq.\ref{eq_ganpf_I1X} into Eq.\ref{eq_ganpf_dfd}, and noting that the right-hand-side term of Eq.\ref{eq_ganpf_I1X} does not depend on $M$, we can take $M\rightarrow \infty$ in Eq.\ref{eq_ganpf_I2} to obtain 
\begin{align}\label{eq_ganpf_dfd2}
    d_{\cF_d}(\mu^*,\tilde \mu_n) = O_{\PP}(n^{-1/2}),
\end{align}
which, together with Eq.\ref{eq_GANapp3}, leads to
\begin{align}\label{eq_GANapp4}
    -\lambda \int_\Omega \log \frac{p_n}{p^*}{\rm d}\hat \mu_n \leq O_{\PP}(n^{-1/2}).
\end{align}
By the triangle inequality and Eq.\ref{eq_GANapp4}, we obtain
\begin{align}\label{eq_GANapp5}
    -\lambda \int_\Omega \log \frac{p_n}{p^*}{\rm d}\mu^* \leq O_{\PP}(n^{-1/2}) + \lambda\left|\int_\Omega \log \frac{p_n}{p^*}{\rm d}(\mu^* - \hat \mu_n)\right|.
\end{align}
It remains to consider 
\begin{align*}
    \left|\int_\Omega \log \frac{p_n}{p^*}{\rm d}(\mu^* - \hat \mu_n)\right|,
\end{align*}
which can be bounded by Lemma 5.11 of \cite{geer2000empirical} again.  

Specifically, since $\cG\in \cW^{m_2}(R_3)$, the entropy number of $\cG$ can be bounded by \cite{adams2003sobolev}
\begin{align*}
    H(u,\cG,\|\cdot\|_P) \leq C\delta^{-d/m_2},
\end{align*}
where $C$ is a constant. Therefore, since $\cG\subset\cW^{m_2}(R_3)$, Lemma 5.11 of \cite{geer2000empirical} gives us
\begin{align}\label{eq_ganpf_ppsd}
    \sup_{p\in \cG}\left|\int_\Omega \log \frac{p}{p^*}{\rm d}(\mu^* - \hat \mu_n)\right| = O_{\PP}(n^{-1/2}),
\end{align}
which, together with Eq.\ref{eq_GANapp5}, gives us
\begin{align*}
    {\rm KL}(\mu^*||\mu_n) = O_{\PP}(\max\{n^{-1/2},\lambda^{-1}n^{-1/2}\}).
\end{align*}
By Eq.\ref{eq_GANapp2}, Eq.\ref{eq_ganpf_dfd2}, and Eq.\ref{eq_ganpf_ppsd}, we have
\begin{align}\label{eq_GANapp6}
    d_{\cF_d}(\mu_n, \mu^*) \leq 2 d_{\cF_d}(\mu^*,\tilde \mu_n) + \lambda\left|\int_\Omega \log \frac{p_n}{p^*}{\rm d}(P - P_n)\right| = O_{\PP}(\max\{n^{-1/2},\lambda n^{-1/2}\}).
\end{align}
Taking $\lambda$ as a constant finishes the proof.

\section{Estimating density of behavior policy using MaxEnt IRL is equivalent to training a GAN with specific likelihood function}\label{appb}

In this section, we prove Proposition \ref{prop1}. In offline reinforcement learning scenario, the distribution of a trajectory $\tau=(s_0,a_0,s_1,a_1,...,s_H,a_H) \in D$ can be written in the following form:
\begin{align}\label{b1}
    P_{\theta}(\tau)=P_{0_{\cD}}(s_0)\prod\limits_{t=0}\limits^{H}{\pi_{\beta}(a_t|s_t)T_{\cD}(s_{t+1}|s_t,a_t)}.
\end{align}
In the training process, $P_{0_{\cD}}(s_0), T_{\cD}(s_{t+1}|s_t,a_t), \forall t \in [0,H]$ are known. Therefore, the uncertainty of the distribution of a trajectory is only related to the probability density of the behavior policy, and we have 
\begin{align}\label{b2}
    P_{\theta}(\tau)=C\prod\limits_{t=0}\limits^{H}{\pi_{\beta}(a_t|s_t)=CL_{\theta}^{\pi_{\beta}}(\tau)},
\end{align}
where $L_{\theta}(\pi_{\beta})$ is the likelihood function of $\pi_{\beta}$ given a trajectory, and $C$ is a constant with respect to dataset $\cD$.

Following \cite{finn2016connection}, in MaxEnt IRL we try to estimate the density of the trajectory by the Boltzmann distribution as
\begin{align}\label{b3}
    p_{\theta}(\tau)=\frac{1}{Z}\exp(-c_{\theta}(\tau)).
\end{align}
In \cite{finn2016connection}, it has been shown that if we estimate $Z$ in the MaxEnt IRL formulation using guided cost learning, and suppose we have a GAN that can give an explicit density of the data, then optimizing the cost function of guided cost learning $\mathcal{L}_{cost}(\theta)$ is equivalent to optimizing the discriminator loss $\mathcal{L}_{discriminator}(D_{\theta})$ in GAN. In the process of estimating $Z$, we also need to train a new sampling distribution $q(\tau)$ and use importance sampling to estimate $Z$. The new sampling policy is optimized by minimizing the KL divergence of $q(\tau)$ and the Boltzmann distribution in Eq.\ref{b3}:
\begin{align}\label{b4}
    \mathcal{L}_{sampler}(q)=\mathbb{E}_{\tau \sim q}[c_{\theta}(\tau)]+\mathbb{E}_{\tau \sim q}[\log(q(\tau))].
\end{align}
The optimal sampling distribution is the demonstration policy (we call behavior policy in offline RL), which is an estimator of the behavior policy. In \cite{finn2016connection} setting, if we assume the sampling policy $q(\tau)$ is just an explicit density of GAN, we will show the generator loss that has a hybrid style as in Eq.\ref{eq4} is equivalent to optimize $\mathcal{L}_{sampler}(q)$. Note that
\begin{align}\label{b4}
    \mathcal{L}_{generator}(q)= &\mathbb{E}_{\tau \sim q}[(\log(1-D_{\tau}))-\log(D_{\tau})]-\lambda\mathbb{E}_{\tau \sim q}[\log(q(\tau))]\nonumber\\
    = &\mathbb{E}_{\tau \sim q}[(\log\frac{q(\tau)}{\Tilde{\mu}(\tau)}-\log\frac{\frac{1}{Z}\exp(-c_{\theta}(\tau))}{\Tilde{\mu}(\tau)}]-\lambda\mathbb{E}_{\tau \sim q}[\log(q(\tau))]\nonumber\\
    = & \mathbb{E}_{\tau \sim q}[(\log(q(\tau))-\log(\frac{1}{Z}\exp(-c_{\theta}(\tau)))]-\lambda\mathbb{E}_{\tau \sim q}[\log(q(\tau))]\nonumber\\
    = & \log(Z)+\mathbb{E}_{\tau \sim q}[c_{\theta}(\tau)]+\mathbb{E}_{\tau \sim q}[(1-\lambda)\log(q(\tau))].
\end{align}
In Eq.\ref{b4}, the term $\log(Z)$ is can be seen as a constant since it is just a normalizing term, so optimizing the generator loss $\mathcal{L}_{generator}(q)$ is equivalent to minimizing the KL divergence between $\Tilde{C} q(\tau)$ and the Boltzmann distribution of the real trajectory density $p_{\theta}(\tau)$. Therefore, if we use a hybrid loss in the generator, the optimization of the generator is still equivalent to the optimization of $\mathcal{L}_{sampler}(q)$ up to a constant $\Tilde{C}$. As $q(\tau) \propto \exp(-c_{\theta}(\tau)) \propto p_{\theta}(\tau)$, and combined with Eq.\ref{b2}, we have 
\begin{align}\label{b5}
   L_{\theta}^{\pi_{\beta}}(\tau)=\frac{1}{C}P_{\theta}(\tau) \propto \exp(-c_{\theta}(\tau)) \propto q(\tau).
\end{align}
If we further denote the sampling policy estimated by GAN as $q^{G}_{\theta}(\tau)$, then we can get the result in Proposition \ref{prop1}.

\section{Details of the performance of the learning policy on the estimated product space of CPED}\label{appc}

In this section, we will give a brief proof of Theorem \ref{thm2}, and show that the learning policy can find the optimal (state, action) w.r.t the training dataset $\cD$.

Suppose the offline replay buffer is $\cD$, with state space $\mathscr{S}=\{\mathcal{S},\mathbb{P}_{\cD}\}$ and action space $\mathscr{A}=\{\mathcal{A},\pi_{\beta}\}$. Suppose the stationary point of the Bellman equation w.r.t the production sample space $\mathscr{S} \times \mathscr{A}$ is $Q^{*}(s^*,a^*)$. We consider two cases.
\begin{enumerate}[label=(\arabic*)]
    \item If $\pi_{\beta}$ is an expert policy, then we have $(s^*,a^*) \in \cD$, $Q(s^*,\pi_{\beta}(s^*))=Q^*$. Thus, the optimal (state, action) pair appears on the seen area of dataset $\cD$.
    \item If $\pi_{\beta}$ is an not an expert policy, then we have $(s^*,a^*) \notin \cD$, $Q(s^*,\pi_{\beta}(s^*))<Q^*$. In this situation, the optimal (state, action) pair appears on the unseen area of dataset $\cD$, which belongs to the product space $\mathscr{S} \times \mathscr{A}$.
\end{enumerate}
CPED mainly focuses on the second scenario, and we give proof of the performance of the learning policy when the optimal stationary point of the Bellman equation is in the unseen area of $\cD$.

Let $\hat{\pi}(s):=\argmax_{a}Q^{*}(s,a)$. We first show that $Q^{*}(s,a)=Q(s,\hat{\pi}^{\Delta}(s))$, where $\hat{\pi}^{\Delta}(s)$ is the probability-controlled learning policy using CPED that is defined on the support of behavior policy $\pi_{\beta}$. Suppose $\hat{\pi}^{\Delta}(s)>0$ on $\mathscr{S} \times \mathscr{A}$ and $|Q(s,a)|\le M, \forall (s,a) \in \mathscr{S} \times \mathscr{A}$. Then we have 
\begin{align}\label{c1}
    Q^{*}(s,a)= & r(s,a)+\gamma\mathbb{E}_{s \sim \mathbb{P}_{\cD}, a' \sim \pi^*}[Q^{*}(s',\pi^{*}(s'))]\nonumber\\
    \le & r(s,a)+\gamma\mathbb{E}_{s \sim \mathbb{P}_{\cD}, a' \sim \hat{\pi}}[Q^{*}(s',a')]\nonumber\\
    = & r(s,a)+\gamma\mathbb{E}_{s \sim \mathbb{P}_{\cD}, a' \sim \hat{\pi}}[Q^{*}(s',a')1_{(a' \in \Delta \pi_{\beta}(s'))}]+\gamma\mathbb{E}_{s \sim \mathbb{P}_{\cD}, a' \sim \hat{\pi}}[Q^{*}(s',a')1_{(a' \notin \Delta \pi_{\beta}(s'))}]\nonumber\\
    \le & r(s,a)+\gamma\mathbb{E}_{s \sim \mathbb{P}_{\cD}, a' \sim \hat{\pi}^{\Delta}}[Q^{*}(s',a')]+\gamma M \mathbb{E}_{s \sim \mathbb{P}_{\cD}, a' \sim \hat{\pi}}[1_{(a' \notin \Delta \pi_{\beta}(s'))}]\nonumber\\
    = &\underbrace{r(s,a)+\gamma\mathbb{E}_{s \sim \mathbb{P}_{\cD}, a' \sim \hat{\pi}^{\Delta}}[Q^{*}(s',a')]}_{I_1} +\underbrace{\gamma M \mathbb{P}((a' \notin \Delta \pi_{\beta}(s')))}_{I_2}
\end{align}
Let action set $\Omega=\{a': \pi_{\beta}(a'|s')=0 , \hat{\pi}(a'|s')>\epsilon, \forall s' \in \mathscr{S} \} $. If the conditional probability space $\pi_{\beta}$ can be estimated accurate, then by CPED, $\mathbb{P}(\Omega)$ is tending to zero. By Theorem \ref{thm1}, when using GAN with hybrid loss to estimate the density of $\pi_{\beta}$, with probability tending to one, we have $\mathbb{P}(\Omega) \to 0$. Then for term $I_2$,
\begin{align}\label{c2}
    I_{2} =  \gamma M \mathbb{P}((a' \notin \Delta \pi_{\beta}(s')))
    \le  \gamma M \mathbb{P}(\Omega) \to 0.
\end{align}
For term $I_1$, based on iteration, we obtain
\begin{align}\label{c3}
    I_{1} =&  r(s,a)+\gamma\mathbb{E}_{s \sim \mathbb{P}_{\cD}, a' \sim \hat{\pi}^{\Delta}}[r(s',a')+\gamma\mathbb{E}_{s \sim \mathbb{P}_{\cD}, a'' \sim \pi^*}[Q^{*}(s'',\pi^{*}(s''))]]\nonumber\\
    \le & r(s,a)+\gamma\mathbb{E}_{s \sim \mathbb{P}_{\cD}, a' \sim \hat{\pi}^{\Delta}}[I_{1}' +\gamma M \mathbb{P}(\Omega)]\nonumber\\
    \le & r(s,a)+\gamma\mathbb{E}_{s \sim \mathbb{P}_{\cD}, a' \sim \hat{\pi}^{\Delta}}[r(s',a')+ \gamma\mathbb{E}_{s \sim \mathbb{P}_{\cD}, a' \sim \hat{\pi}^{\Delta}} [I_{1}^{''} +\gamma M \mathbb{P}(\Omega)]+\gamma M \mathbb{P}(\Omega)]\nonumber\\
    \le & \mathbb{E}[r(s,a)+\gamma r(s',\hat{\pi}^{\Delta}(s'))+\gamma^2 r(s'',\hat{\pi}^{\Delta}(s''))+\dots]+(\gamma+\gamma^2+\gamma^3+\dots)M \mathbb{P}(\Omega)\nonumber\\
    = & Q(s,\hat{\pi}^{\Delta}(s))+\frac{\gamma}{1-\gamma}M\mathbb{P}(\Omega).
\end{align}
Since $\mathbb{P}(\Omega) \to 0$, combining Eq.\ref{c1} and Eq.\ref{c3} yields
\begin{align}\label{c4}
    Q^{*}(s,a) \le & I_{1}+I_{2}\nonumber\\
    \le & Q(s,\hat{\pi}^{\Delta}(s))+\frac{\gamma}{1-\gamma}M\mathbb{P}(\Omega)+\gamma M \mathbb{P}(\Omega) \to Q(s,\hat{\pi}^{\Delta}(s))
\end{align}
On the other hand, obviously $Q^{*}(s,a) \ge Q(s,\hat{\pi}^{\Delta}(s))$, a.s..

Hence, with probability tending to one, $Q^{*}(s,a)$ is close to $Q(s,\hat{\pi}^{\Delta}(s))$. 

Then the learning policy $\hat{\pi}^{\Delta}$ trained by CPED is close to the optimal policy in product space $\mathscr{S} \times \mathscr{A}$ with probability tending to one.

\section{Details of the convergence learning policy of CPED when using policy iteration}\label{appd}

In this section, we will give a brief proof of Theorem \ref{thm3}, and show the convergence of the learning policy when using policy iteration to update the learning policy. The whole proof is divided into two parts. First, we show the monotonic improvement of Q function of the iterated policy by CPED. Then we give the convergence of the iterated value function.

\subsection{Monotonic improvement of Q function $Q^{\pi_{k+1}^{\Delta}}(s,a) \ge Q^{\pi_{k}^{\Delta}}(s,a)$}

\begin{proof}\label{prof2}

Suppose we use iteration to improve our policy in each training step by 
\begin{align}\label{d1}
\pi_{k+1}:=\argmax_{a}Q^{\pi_{k}}(s,a),\forall s \in \mathscr{S}
\end{align}
Then at iteration $k+1$, we have
\begin{align}\label{d2}
Q^{\pi_{k+1}}(s,a)- Q^{\pi_{k}}(s,a)=& \gamma\mathbb{E}_{s' \sim \mathbb{P}_{\cD}}[Q^{\pi_{k+1}}(s',\max_{a'_{k+1}}\pi_{k+1}(s'))-Q^{\pi_{k}}(s',\max_{a'_{k}}\pi_{k}(s'))]\nonumber\\
=& \gamma\mathbb{E}_{s' \sim \mathbb{P}_{\cD}}[Q^{\pi_{k+1}}(s',\max_{a'_{k+1}}\pi_{k+1}(s'))-Q^{\pi_{k}}(s',\max_{a'_{k+1}}\pi_{k+1}(s'))+\nonumber\\
& \underbrace{Q^{\pi_{k}}(s',\max_{a'_{k+1}}\pi_{k+1}(s'))-Q^{\pi_{k}}(s',\max_{a'_{k}}\pi_{k}(s'))}_{\ge 0, \text{by definition of policy iteration}}]\nonumber\\
\ge & \gamma\mathbb{E}_{s' \sim \mathbb{P}_{\cD}}[Q^{\pi_{k+1}}(s',\max_{a'_{k+1}}\pi_{k+1}(s'))-Q^{\pi_{k}}(s',\max_{a'_{k+1}}\pi_{k+1}(s'))]\nonumber\\
\ge & \gamma\mathbb{E}_{s' \sim \mathbb{P}_{\cD}, a'_{k+1} \sim \pi_{k+1}}[Q^{\pi_{k+1}}(s',a'_{k+1})-Q^{\pi_{k}}(s',a'_{k+1})]\nonumber\\
= & \underbrace{\gamma\mathbb{E}_{s' \sim \mathbb{P}_{\cD}, a'_{k+1} \sim \pi_{k+1}}[(Q^{\pi_{k+1}}(s',a'_{k+1})-Q^{\pi_{k}}(s',a'_{k+1}))1_{(a'_{k+1} \in \Delta \pi_{\beta}(s))}]}_{I_1}+\nonumber\\
& \underbrace{\gamma\mathbb{E}_{s' \sim \mathbb{P}_{\cD}, a'_{k+1} \sim \pi_{k+1}}[(Q^{\pi_{k+1}}(s',a'_{k+1})-Q^{\pi_{k}}(s',a'_{k+1}))1_{(a'_{k+1} \notin \Delta \pi_{\beta}(s))}]}_{I_2}.
\end{align}

For term ${I_2}$, let $X_{k+1}=(Q^{\pi_{k+1}}(s',a'_{k+1})-Q^{\pi_{k}}(s',a'_{k+1}))1_{(a'_{k+1} \notin \Delta \pi_{\beta}(s))}$. Then it holds that
\begin{enumerate}[label=(\arabic*)]
\item $X_{k+1} \to 0$ in probability, as $1_{(a'_{k+1} \notin \Delta \pi_{\beta}(s))} \to 0$ by policy control defined in CPED.
\item $X_{k+1}$ is bounded by some constant.
\end{enumerate}
So based on the Dominated Convergence Theorem, $\mathbb{E}X_{k+1} \to 0$.

For term ${I_1}$, we have
\begin{align}\label{d3}
I_1=& \gamma\mathbb{E}_{s' \sim \mathbb{P}_{\cD}, a'_{k+1} \sim \pi_{k+1}}[(Q^{\pi_{k+1}}(s',a'_{k+1})-Q^{\pi_{k}}(s',a'_{k+1}))1_{(a'_{k+1} \in \Delta \pi_{\beta}(s))}]\\
=& \gamma\mathbb{E}_{s' \sim \mathbb{P}_{\cD}, a'_{k+1} \sim \pi_{k+1}^{\Delta}}[Q^{\pi_{k+1}}(s',a'_{k+1})-Q^{\pi_{k}}(s',a'_{k+1})].
\end{align}
By iteration,
\begin{align}\label{d4}
& (Q^{\pi_{k+1}}(s',a'_{k+1})-Q^{\pi_{k}}(s',a'_{k+1}))1_{(a'_{k+1} \in \Delta \pi_{\beta}(s))}\nonumber\\
\ge & \gamma\mathbb{E}_{s'' \sim \mathbb{P}_{\cD}, a''_{k+1} \sim \pi_{k+1}^{\Delta}}[Q^{\pi_{k+1}}(s'',a''_{k+1})-Q^{\pi_{k}}(s'',a''_{k+1})] + \gamma I_2.
\end{align}
So we have
\begin{align}\label{d5}
Q^{\pi_{k+1}}(s,a)- Q^{\pi_{k}}(s,a) \ge  \gamma^{\infty}+\frac{\gamma}{1-\gamma} I_2 \to 0 \quad when \quad a \in \Delta \pi_{\beta}.
\end{align}
Therefore, for sufficiently large $k$, $Q^{\pi_{k+1}^{\Delta}}(s,a) \ge Q^{\pi_{k}^{\Delta}}(s,a)$. Furthermore, we have $V^{\pi_{k+1}^{\Delta}}(s) \ge V^{\pi_{k}^{\Delta}}(s)$. 
\end{proof}

\subsection{Convergence of the iterated value function$\Vert V^{\hat{\pi}^{\Delta}_{k+1}}-V^* \Vert_{\infty} \le \gamma^{k+1} \Vert V^{\hat{\pi}^{\Delta}_{0}}-V^* \Vert_{\infty}$}

\begin{proof}\label{prof3}
Direct computation shows that
\begin{align}\label{d6}
V^*(s)-V^{\hat{\pi}^{\Delta}_{k+1}}(s)=& \max_{a}[r(s,a)+\gamma\mathbb{E}_{s' \sim \mathbb{P}_{\cD}}V^*(s')]-[r(s,\hat{\pi}^{\Delta}_{k+1}(s))+\gamma\mathbb{E}_{s' \sim \mathbb{P}_{\cD}}V^{\hat{\pi}^{\Delta}_{k+1}}(s')]\nonumber\\
\le & \max_{a}[r(s,a)+\gamma\mathbb{E}_{s' \sim \mathbb{P}_{\cD}}V^*(s')]-[r(s,\hat{\pi}^{\Delta}_{k+1}(s))+\gamma\mathbb{E}_{s' \sim \mathbb{P}_{\cD}}V^{\hat{\pi}^{\Delta}_{k}}(s')\nonumber\\
=& \max_{a}[r(s,a)+\gamma\mathbb{E}_{s' \sim \mathbb{P}_{\cD}}V^*(s')]-\max_{a}[r(s,a)+\gamma\mathbb{E}_{s' \sim \mathbb{P}_{\cD}}V^{\hat{\pi}^{\Delta}_{k}}(s')]\nonumber\\
\le & \max_{a}(r(s,a)+\gamma\mathbb{E}_{s' \sim \mathbb{P}_{\cD}}V^*(s')-(r(s,a)+\gamma\mathbb{E}_{s' \sim \mathbb{P}_{\cD}}V^{\hat{\pi}^{\Delta}_{k}}(s')))\nonumber\\
\le & \gamma \Vert V^*-V^{\hat{\pi}^{\Delta}_{k}}\Vert_{\infty}
\le \gamma^2 \Vert V^*-V^{\hat{\pi}^{\Delta}_{k-1}}\Vert_{\infty} \le \dots \le \gamma^{k+1} \Vert V^{\hat{\pi}^{\Delta}_{0}}-V^* \Vert_{\infty}.
\end{align}
Since Eq.\ref{d6} holds for $\forall s \in \mathscr{S}$, we have 
\begin{align*}
\Vert V^{\hat{\pi}^{\Delta}_{k+1}}-V^* \Vert_{\infty} \le \gamma^{k+1} \Vert V^{\hat{\pi}^{\Delta}_{0}}-V^* \Vert_{\infty}.
\end{align*}
This finishes the proof.
\end{proof}

\section{Experiment Details and More Results}\label{appe}

\subsection{The Mujoco and AntMaze tasks}
\label{dataset}

\begin{itemize}[leftmargin=*]

    \item   
  \textbf{Gym-MuJoCo Task.} The Gym-MuJoCo is a commonly used benchmark for offline RL task. Each Gym-MuJoCo task contains three types of static offline datasets, generated by different behavior policies. The three offline datasets include:
\textbf{i)} Random dataset originates from random policies and contains  the least valuable information 
\textbf{ii)} The medium-quality dataset is generated by a partially trained policy (which performs about 1/3 as well as an expert policy).The medium-quality dataset contains three types of mixture: medium, medium-replay,medium-expert.
\textbf{iii)} The expert dataset is produced by a policy trained to completion with SAC.

In offline RL, the datasets generated by the medium-quality policy are better considered since we are confronted with such dataset in most practical scenarios. Therefore, in our experiments, we consider the datasets from the medium-quality policy.

    \item \textbf{AntMaze Task.}
The AntMaze task is a challenging navigation task that requires a combination of different sub-optimal trajectories to find the optimal path. According to different task levels, the AntMaze task is divided into umaze, medium, and large types.
\end{itemize}

\subsection{Implementation Details and Discussion}
\label{implemention}

Implementation of CPED algorithm contains Flow-GAN training and Actor-Critic training.

\textbf{Flow-GAN training.} The generator of GAN follows the NICE \cite{Dinh2015} architecture of the flow model, and the discriminator is a 4-layer MLPs. The entire code of training Flow-GAN is based on the implementation of FlowGAN\cite{Grover2018}\footnote{ https://github.com/ermongroup/flow-gan}.  FlowGAN\cite{Grover2018} is intended to deal with image data, while the D4RL dataset does not contain image features. So in our training, we replace the image data-specific network architecture(such as convolution layers, Residual block) with fully connected layers. The hyperparameters used in Flow-GAN is shown in Table \ref{tab3}.

\begin{table*}[t]
\caption{The hyperparameter used in Flow-GAN training.}
\label{tab3}
\vskip 0.15in
\begin{center}
\begin{small}
\begin{tabular}{lll}
\toprule
 & Hyperparameter & Value\\
\midrule
  &Optimizer & Adam  \\
 &Batch size & 256\\
    &Gradient penalty for WGAN-GP\cite{gulrajani2017improved} & 0.5  \\
  Shared parameter  &Training ratio of the generator and discriminator  & 5:1 for DCGAN\cite{radford2015unsupervised}   \\
    &  & 1:5 for WGAN-GP   \\
     &Maximium Likelihood Estimation(MLE) adjustment weight  & 1 \\
     &  Activation in the hidden layer&  LeakyRelu\\
     
\midrule
      &Learning rate & 1e-4  \\
    &Latent layer of m network in NICE\cite{Dinh2015} & 3 \\
Generator    &Latent dim & 750 \\
   &Hidden Layer & 4 \\
   &Output activation & LeakyRelu \\
\midrule
      &Learning rate & 1e-5 for DCGAN \\
      & & 1e-4 for WGAN-GP \\
   &Latent dim & [2 $\times$ (action dim +state dim),\\
   &&4 $\times$ (action dim +state dim)] \\
Discriminator   &Hidden Layer & 2 \\
   &Normalization & Batch Norm for DCGAN \\
   & & Layer Norm for WGAN-GP \\
   &Output activation & Sigmoid for DCGAN\\
   & & Identity for WGAN-GP\\
   & Dropout rate & 0.2 for DCGAN\\
   &  & 0.0 for WGAN-GP\\

\bottomrule
\end{tabular}
\end{small}
\end{center}
\vskip -0.1in
\end{table*}

\textbf{Actor-Critic training.} The Actor-Critic training implementation follows TD3\cite{fujimoto2021minimalist}, as SPOT\cite{wu2022supported} suggested TD3 architecture performs better than SAC in offline RL tasks. Additionally, the reward in AntMaze tasks is centered following the implementation of CQL/IQL. When training CPED in practice, the Flow-GAN updates first after initializing the CPED. The policy and $Q$ networks start to update when Flow-GAN has trained 20 thousand steps, and a relatively reliable density estimator is provided. Afterward, the Actor-Critic begins to train simultaneously. To prevent overfitting, the Flow-GAN stops training when it becomes stable.The hyperparameters used in Actor-Critic training are shown in Table \ref{tab4}.

\begin{table*}[t]
\caption{The hyperparameters used in Actor-Critic training.}
\label{tab4}
\vskip 0.15in
\begin{center}
\begin{small}
\begin{tabular}{lll}
\toprule
 & Hyperparameter & Value\\
\midrule
  &Optimizer & Adam  \\
 &Batch size & 256\\
    &Actor learning rate  & 3e-4  \\
  TD3  &Critic learning rate  & 3e-4   \\
     &Discount factor  & 0.99   \\
     &Number of iterations  & 1e6\\
     &Target update rate  $\tau$ & 0.005\\
     &Policy noise&  0.2\\
     &Policy noise clipping&  0.5\\
     &Policy update frequency&2\\
     
\midrule
      &Actor hidden dim & 256  \\
Architecture     &Actor layers & 3 \\
   &Critic hidden dim & 256 \\
   &Critic layers & 3 \\
\midrule
 CPED     &Time varying $\alpha$ & see in Figure \ref{fig3}(b)-(c) \\

\bottomrule
\end{tabular}
\end{small}
\end{center}
\vskip -0.1in
\end{table*}

\textbf{Hyperparameter tuning of the time(epoch) varying $\alpha$.} The key element in  the time-varying setting of policy constraint parameter $\alpha$ is to find the task's strongest and weakest $\alpha$ values. We follow the pattern we observed in automatic learning of $\alpha$ in BEAR\cite{Kumar2019} and set the strongest and weakest $\alpha$ values for different offline RL tasks. We find that after 200-300 thousand steps of learning with the strongest policy constraint, we get a relatively strong learning policy. Then we decrease $\alpha$ to the weakest value. Indeed, this decreasing process is robust to small changes of  $\alpha$. The settings in Figure \ref{fig3}(b) and \ref{fig3}(c) are generally applicable for most tasks in Mujoco task and AntMaze task.

\subsection{Training Curve}

\begin{figure}[!h]
\vskip 0.2in
\begin{center}
\includegraphics[width=\linewidth]{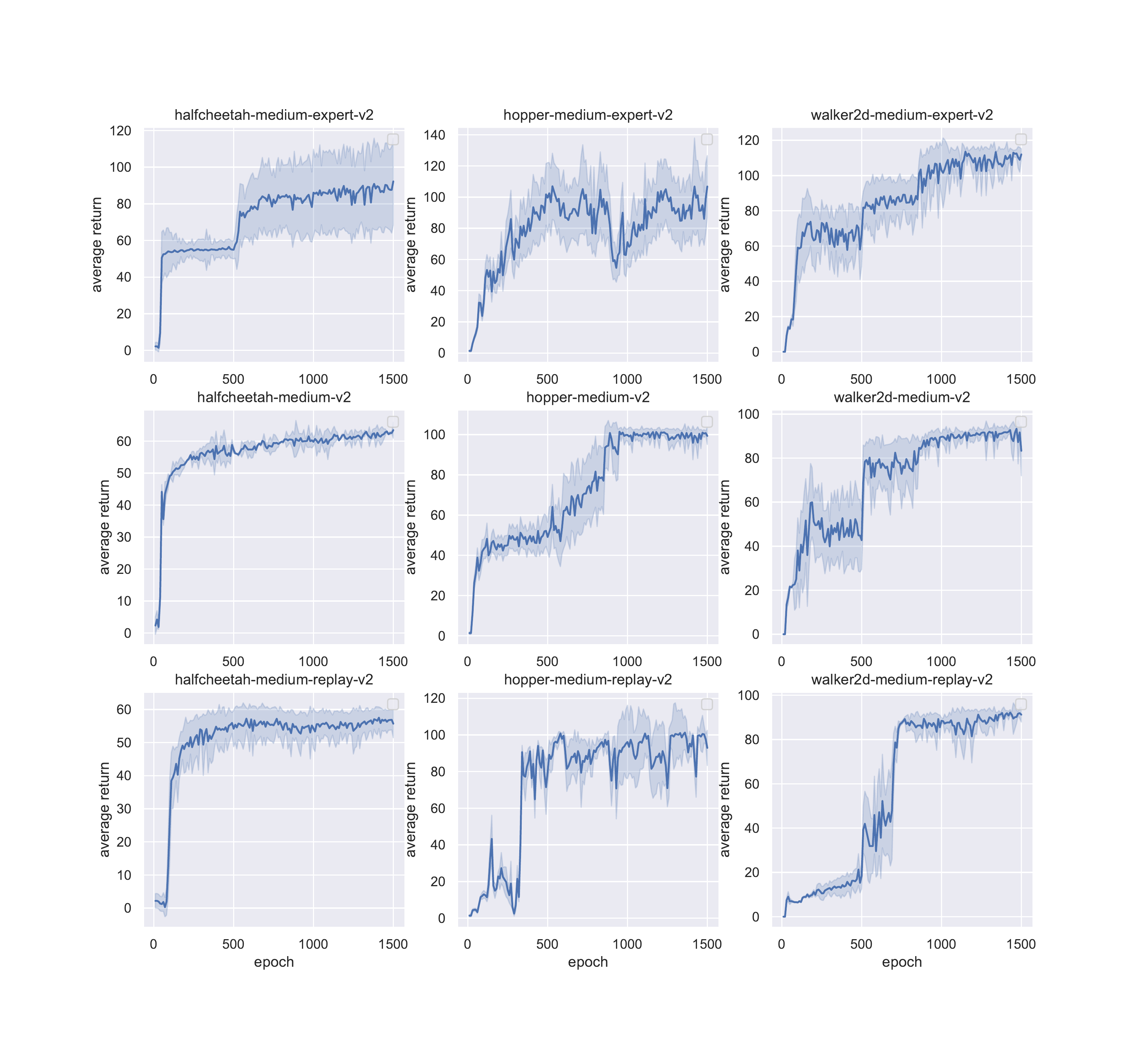}
\caption{
Training curve of different Mujoco Tasks. All results are averaged across 5 random seeds. Each epoch contains 1000 training steps.
}
\label{fig5}
\end{center}
\vskip -0.2in
\end{figure}

\begin{figure}[!h]
\vskip 0.2in
\begin{center}
\includegraphics[width=\linewidth]{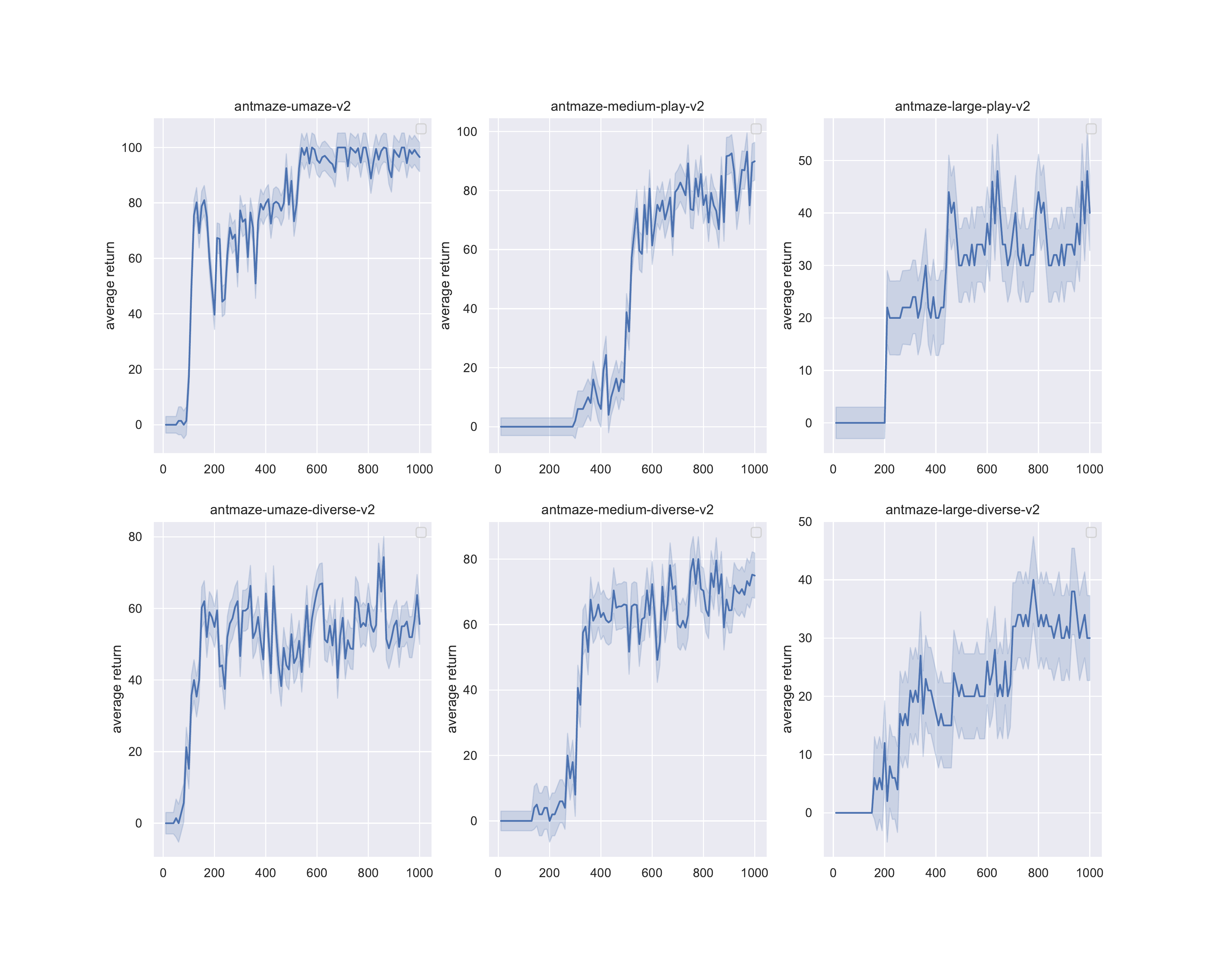}
\caption{
Training curve of different Antmaze Tasks. All results are averaged across 5 random seeds. Each epoch contains 1000 training steps.
}
\label{fig6}
\end{center}
\vskip -0.2in
\end{figure}

The training curve of Mujoco tasks is shown in Figure \ref{fig5}. The training curve of AntMaze tasks is shown in Figure \ref{fig6}.

\subsection{Target Q function during training}

\begin{figure}[!h]
\vskip 0.2in
\begin{center}
\includegraphics[width=\linewidth]{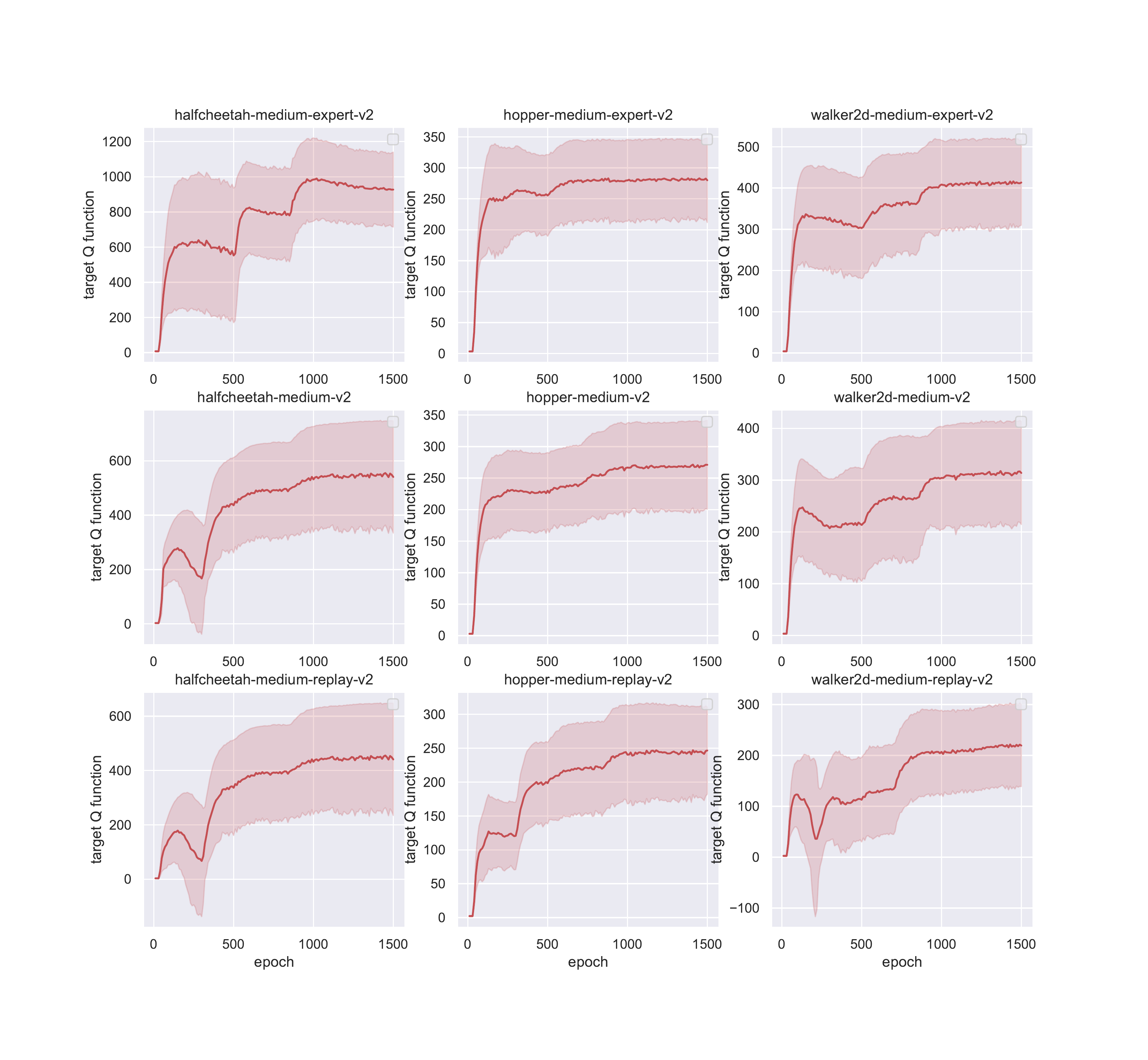}
\caption{
Target Q function of different Mujoco Tasks. All results are averaged across 5 random seeds. Each epoch contains 1000 training steps.
}
\label{fig7}
\end{center}
\vskip -0.2in
\end{figure}

We show the target Q function of different Mujoco Tasks in Figure \ref{fig7} and give an experimental analysis of the policy constraint method. The target Q function of the AntMaze task is quite small due to the sparse reward, and we only show the target Q function of the Mujoco Task. In the offline RL scenario, the target Q function in the Bellman equation is vulnerable to OOD points. For OOD points, the target Q function are overestimated, and the Q value could be very large when the extrapolation errors are accumulated continuously. As a consequence, the learning policy fails to predict a reasonable action for most cases. CQL\cite{Kumar2020} discusses that policy control alone can not prevent the target Q function from being overestimated on OOD points. The experiment shows if we can get an explicit and relatively accurate density function of the behavior policy, we can get a stable and meaningful target Q function.

\subsection{Ablations}\label{ablationinapp}

\begin{figure}[!h]
\vskip 0.2in
\begin{center}
\includegraphics[width=\linewidth]{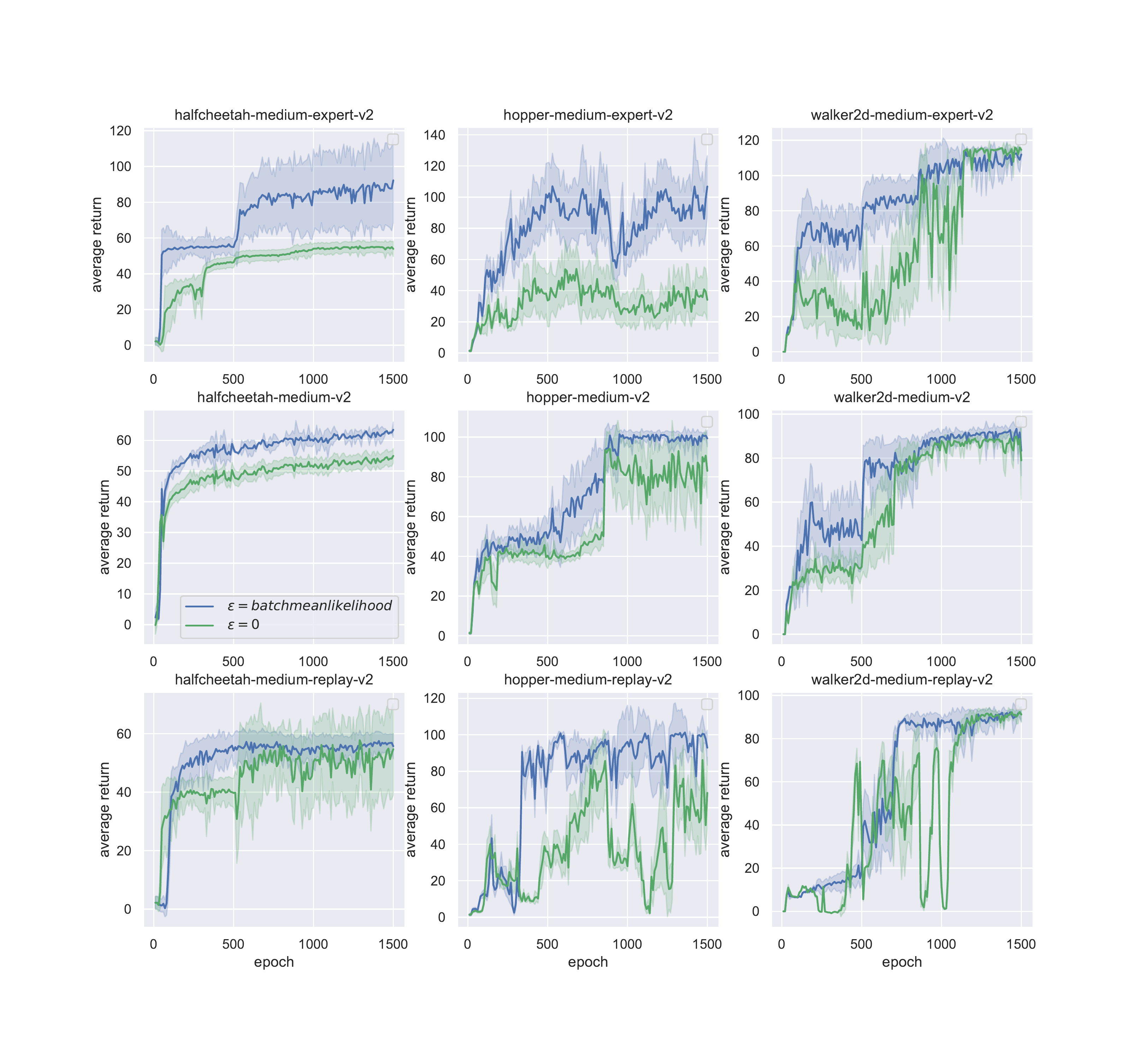}
\caption{
Training curve of different Mujoco Tasks when using different likelihood threshold $\epsilon$. All results are averaged across 5 random seeds. Each epoch contains 1000 training steps. The green line denotes the training curve under $\epsilon = 0$, and the blue line denotes the training curve under our batch mean likelihood setting.
}
\label{fig8}
\end{center}
\vskip -0.2in
\end{figure}

In the ablation study, we compare the performances of two different settings of $\epsilon$ in Eq.\ref{eq7}, in which $\epsilon$ is set to 0 (traditional quantile setting ) and batch mean likelihood (our setting), respectively\cite{Kumar2019,wu2022supported}.  The learning curves of these two settings are shown in Figure \ref{fig8}. We only show the performance of the Mujoco task as AntMaze tasks are very unstable in the ablation study. For most Mujoco tasks, the batch mean likelihood setting for $\epsilon$ achieves higher normalized return, while the learning cure of the traditional quantile setting is quite volatile.

\begin{figure}[!h]
\vskip 0.2in
\begin{center}
\includegraphics[width=\linewidth]{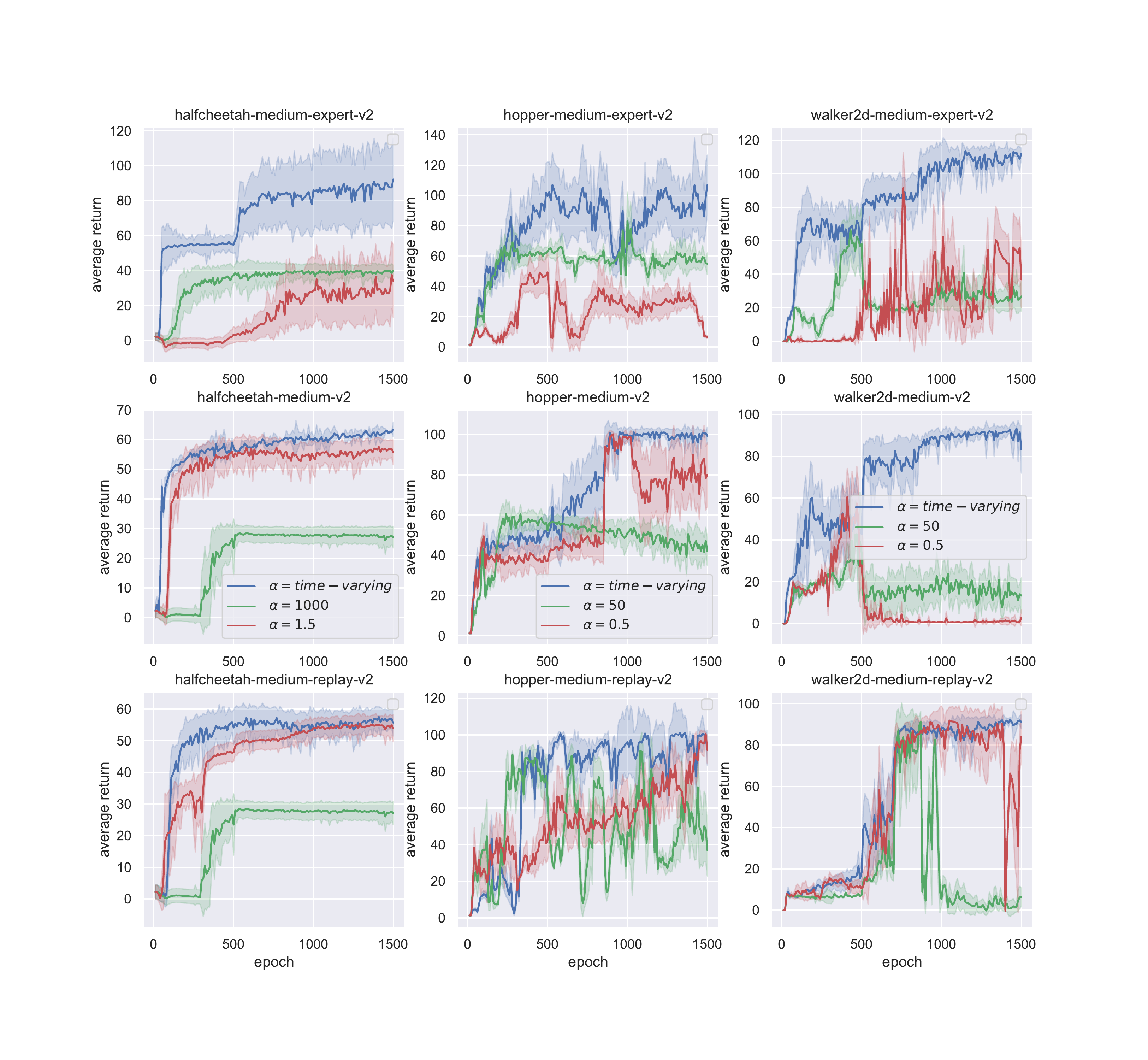}
\caption{
Training curve of different Mujoco Tasks when using different constraint threshold $\alpha$. All results are averaged across 5 random seeds. Each epoch contains 1000 training steps. For each task (halfcheetah/hopper/walker2d), the selected $\alpha$ in the experiment remains the same under three types of dataset (medium-expert-v2/medium-v2/medium-replay-v2). 
}
\label{fig9}
\end{center}
\vskip -0.2in
\end{figure}

We further analyze the time-varying constraint setting we used in CPED by comparing its performance with that of the constant constraint setting. To better show the differences between two kinds of constraint settings on $\alpha$, we take the maximum and minimum values of the time-varying constraint as the constant constraint value. The learning curves of these two settings are shown in Figure \ref{fig9}.  We only show the Mujoco task performance as AntMaze tasks are unstable. For most Mujoco tasks, the time-varying constraint setting of $\alpha$ outperforms the constant setting with a large margin.

\begin{figure}[!h]
\vskip 0.2in
\begin{center}
\includegraphics[width=\linewidth]{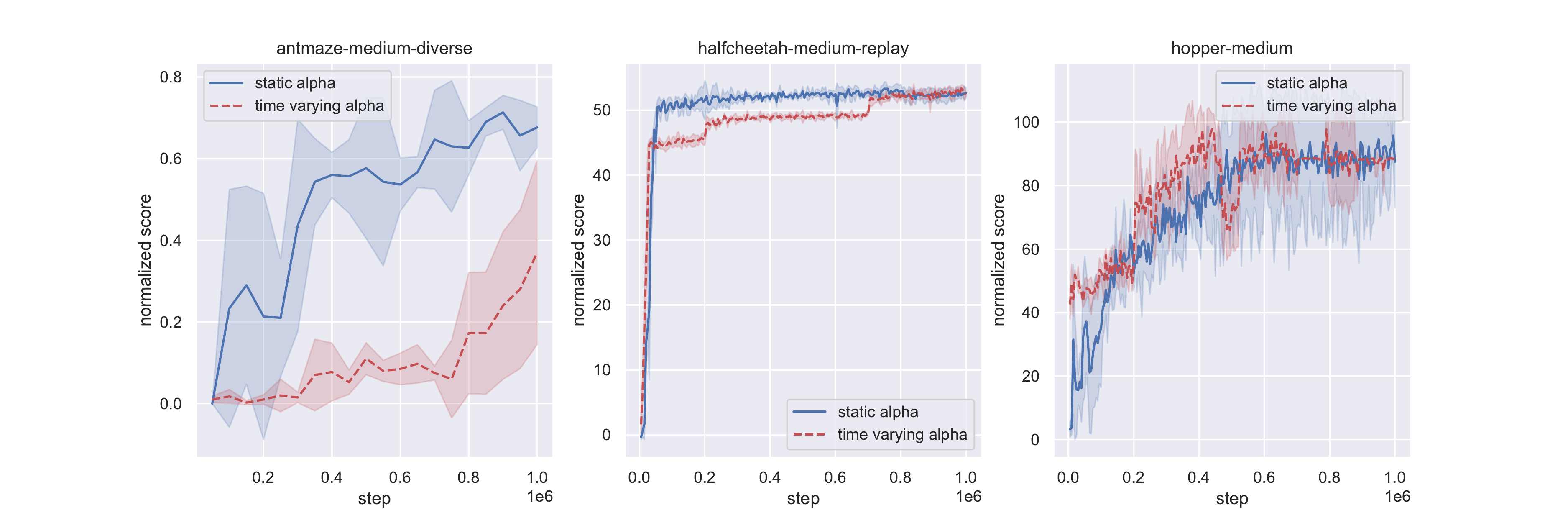}
\caption{
Training curve of three randomly chosen tasks for spot when using constant $\alpha$ and the time-varying $\alpha$ setting. All results are averaged across 5 random seeds. Each epoch contains 1000 training steps. The blue line denotes the training curve using SPOT's constant $\alpha$ setting, and the red line denotes the training curve using time-varying $\alpha$ setting.
}
\label{fig10}
\end{center}
\vskip -0.2in
\end{figure}

We also used the time-varying constraint trick for Spot and randomly selected three tasks to compare the performance difference between Spot using its own constant alpha and the time-varying alpha setting (Figure \ref{fig10}). We can see that the performances with time-varying $\alpha$ are close or even inferior to those with constant $\alpha$ setting, indicating the trick of varying does not result in too much benefit.

\subsection{Density estimation accuracy by Flow-GAN and VAE: a toy example}\label{densityinapp}

To assess the accuracy of Flow-GAN for distribution estimation, this section conducts a straightforward toy experiment. The experiment involves a comparison between the performance of Flow-GAN and VAE in learning Gaussian mixture distributions and generating samples. Specifically, two Gaussian distribution settings were employed in the experiment:

\begin{itemize}
\item  We gather approximately $12800000$ data points from a multivariate Gaussian distribution with a mean of $\mu=[1,9]$ and covariance matrix of $\Sigma=[[1,0],[0,1]]$.
\item  We gather a similar amount (with setting 1) data points from a Gaussian mixture distribution, in which data are randomly drawn from two independent Gaussian distributions ($\mu_1=[1,1]$ and $\mu_2=[9,9]$, $\Sigma_1=\Sigma_2=[[1,0],[0,1]]$) with equal probability.
\end{itemize}

We then employ both VAE and Flow-GAN to generate samples and calculate the mean of log-likelihood. Table \ref{tab5} demonstrates that Flow-GAN outperforms VAE significantly in approximating the original data distribution. So, as many previous studies \cite{wang2022diffusion,yang2022behavior,chen2022offline,ghasemipour2021emaq} have shown, VAE has shortcomings in estimating complex distributions, especially multimodal distributions or complex behavior policy distributions.Flow-GAN has a stronger advantage in learning complex distributions due to the use of MLE and GAN-based adversary loss.

\begin{table*}[t]
\caption{The mean of log-likelihood of the generated sample by Flow-GAN and VAE.}
\label{tab5}
\vskip 0.15in
\begin{center}
\begin{small}
\begin{tabular}{llll}
\toprule
Setting & Ground Truth & VAE & Flow-GAN \\
\midrule
setting1  &-2.89 & -140.83 & -24.18 \\
setting2 &-2.78 & -50.05 & -6.12\\
\bottomrule
\end{tabular}
\end{small}
\end{center}
\vskip -0.1in
\end{table*}

\end{document}